%% file: src_main.tex
\documentclass[]{acmsiggraph}

\pdfoutput=1

\TOGonlineid{}
\TOGvolume{}
\TOGnumber{}
\TOGarticleDOI{}
\TOGprojectURL{}
\TOGvideoURL{}
\TOGdataURL{}
\TOGcodeURL{}

\usepackage{graphicx}
\usepackage{lineno}
\usepackage{float}
\usepackage{braket}
\usepackage{amssymb}
\usepackage{mathrsfs}
\usepackage{enumitem}

\usepackage{caption}
\usepackage{subcaption}

\usepackage{color}
\usepackage{xspace}

\hypersetup{draft}

\title{Recent Advances in Transient Imaging: \\
A Computer Graphics and Vision Perspective}

\author{Adrian Jarabo \qquad Belen Masia \qquad Julio Marco \qquad Diego Gutierrez
\\
Universidad de Zaragoza, I3A 
}
\pdfauthor{Adrian Jarabo et al.}


\input{src_defines}

\begin{document}

\teaser{
\centering
\includegraphics[width=.95\textwidth]{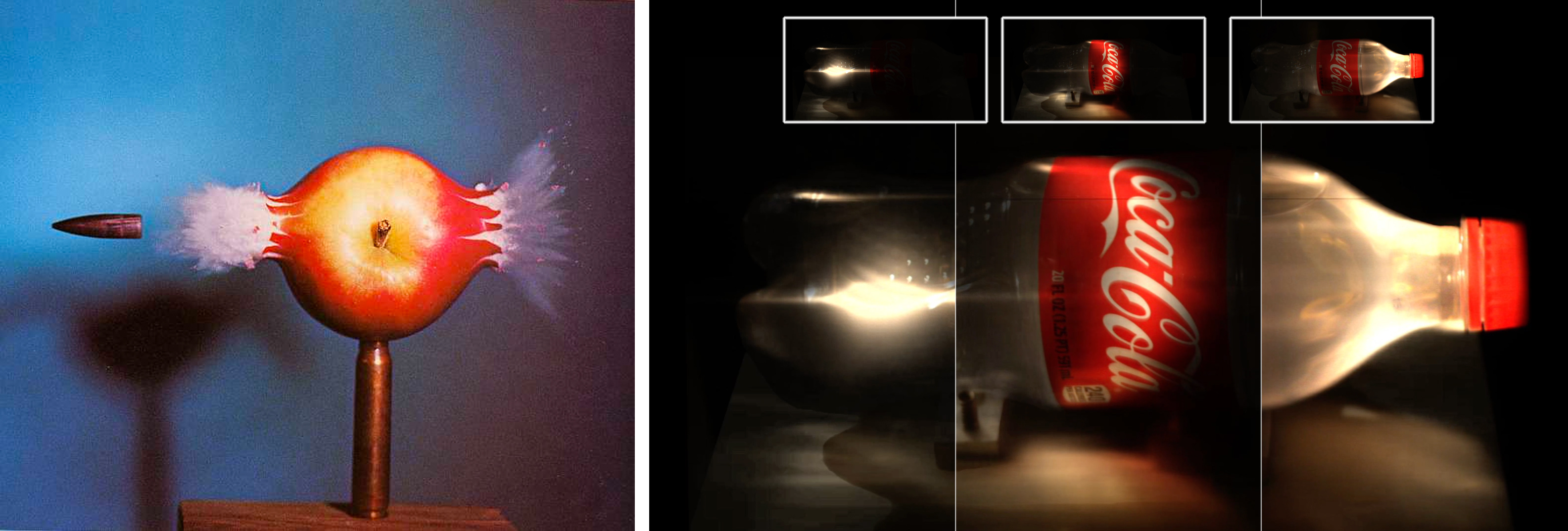}
%\vspace{-.2cm} 
\caption{\textbf{Left:} In 1964, Harold Edgerton captured the iconic \emph{Bullet Through Apple} image (\copyright~MIT Museum). The bullet traveled at about 850 m/s, which translated into an exposure of approximately 4-10 millionth of a second. \textbf{Right:} Almost 50 years later, the femto-photography technique was introduced \protect \cite{Velten2013}, capable of capturing \emph{light in motion}, with an effective exposure time of one \emph{trillionth} of a second. The large split image is a composite of the three complete frames shown in the insets. The complete videos of this and other scenes can be downloaded from \textit{http://giga.cps.unizar.es/~ajarabo/pubs/femtoSIG2013/}. }  
%\vspace{-.4cm} 
\label{fig:teaser}
 }
 
\maketitle

\input{src00_abstract}

%\begin{CRcatlist}
%  \CRcat{I.3.7}{Computer Graphics}{Three-Dimensional Graphics and Realism}{Raytracing}
%\end{CRcatlist}

%\keywordlist

%\TOGlinkslist

%\copyrightspace

%%%%%%%%%%%%%%%%%%%%%%%%%%%%%%%%%%%%%%%%%%%%%%%%%%%
%\linenumbers
\input{src10_intro}

\input{src20_capture}

\input{src30_forward_models}

\input{src40_applications}
\input{src50_simulation}

%\input{src50_simulation_diego}
%\input{src50_visualization_and_analysis}
\input{src60_discussion}
%\input{src8_future_work}
\input{srcZ_acks}
\bibliographystyle{acmsiggraph}
\bibliography{src_bib}

%\clearpage
%\appendix
%\input{srcA_transient_ppm}
%\input{srcB_time_sampling_derivations}

\end{document}

%% file: src_defines.tex
\definecolor{red}{rgb}{0.8,0,0}
\definecolor{darkred}{rgb}{0.6,0,0}
\definecolor{green}{rgb}{0.0,0.5,0}
\definecolor{blue}{rgb}{0,0,0.75}
\definecolor{orange}{rgb}{1,0.6,0.2}
\definecolor{purple}{rgb}{0.7,0.0,0.7}
\definecolor{cyan}{rgb}{0.0,0.7,0.7}

\newcommand{\Fig}[1]{Figure~\ref{fig:#1}}

\newcommand{\Tab}[1]{Table~\ref{tab:#1}}
\newcommand{\Eq}[1]{Equation~\eqref{eq:#1}}

\newcommand{\Sec}[1]{Section~\ref{sec:#1}}
\newcommand{\Secs}[2]{Sections~\ref{sec:#1} and \ref{sec:#2}}

\newcommand{\ToF}{ToF\xspace}			
\newcommand{\AMCW}{ToF\xspace}

\newcommand{\vect}[1]{\mathbf{#1}}           % vector notation
\newcommand{\transient}[1]{#1}

\newcommand{\diff}[1]{d{#1}}

\newcommand{\sImage}{\vect{i}} 
\newcommand{\sLight}{\vect{p}} 
\newcommand{\sSensorResp}{\vect{s}} 
\newcommand{\sTransport}{\vect{T}}
\newcommand{\sTransImage}{\transient{\sImage}} 
\newcommand{\sTransLight}{\transient{\sLight}} 
\newcommand{\sTransTransport}{\transient{\sTransport}}
\newcommand{\sTransSensor}{\transient{\sSensorResp}}
\newcommand{\sImageSize}{I} 
\newcommand{\sLightSize}{P} 
 
\newcommand{\sTransportPixel}{\vect{t}}
\newcommand{\sTransTransportPixel}{\transient{\sTransportPixel}}

\newcommand{\sIndirectDiffuse}{\transient{\vect{l}}}
\newcommand{\sSSS}{\transient{\vect{v}}}

		\newcommand{\approptoinn}[2]{\mathrel{\vcenter{\offinterlineskip\halign{
				\hfil$##$\cr#1\propto\cr\noalign{\kern2pt}#1\sim\cr\noalign{\kern-2pt}
		}}}}

		\newcommand{\sSpeedOfLight}{c}

		\newcommand{\ToFSensorFreq}{f_{\omega_R}}
		\newcommand{\ToFSensorPeriod}{T_{\omega_R}}
		\newcommand{\ToFEmitterFreq}{g_{\omega_T}}

		\newcommand{\Order}[1]{O(#1)}		

%% file: src00_abstract.tex
%!TEX root = src_main.tex
\begin{abstract}
\end{abstract}

Transient imaging has recently made a huge impact in the computer graphics and computer vision fields. By capturing, reconstructing, or simulating light transport at extreme temporal resolutions, researchers have proposed novel techniques to show movies of light in motion, see around corners, detect objects in highly-scattering media, or infer material properties from a distance, to name a few. The key idea is to leverage the wealth of information in the temporal domain at the pico or nanosecond resolution, information usually lost during the capture-time temporal integration. This paper presents recent advances in this field of transient imaging from a graphics and vision perspective, including capture techniques, analysis, applications and simulation.

%% file: src10_intro.tex
	%!TEX root = src_main.tex
\section{Introduction}
\label{sec:intro}

In 1964, MIT professor Harold Edgerton produced the now-iconic \textit{Bullet Through Apple} photograph (see Figure~\ref{fig:teaser}, left). His work represented an unprecedented  effort to photograph events too fast to be captured with traditional techniques. He invented a new stroboscopic flash light (which he termed the stroboscope), which would shine for about 10 microseconds: bright enough, and short enough, to effectively freeze the world and capture ultrafast events such as the bullet bursting through the apple, a splash of a drop of milk, or the flapping wings of a hummingbird. Almost fifty years later, inspired by these images, the technique known as \textit{femto-photography}~\cite{Velten2013} was introduced; it took Edgerton's vision to a whole new level, by allowing to capture movies of light in motion, as it traversed a macroscopic scene (Figure~\ref{fig:teaser}, right).

This fifty-year span provides a clear example of the progress in ultrafast imaging. Many techniques have appeared in the last few years, some inspired by femto-photography, others following completely different approaches. They share the common goal of trying to make visible the invisible: Whether it is due to the speed of the event being captured, to the presence of scattering media, to the lack of photons, or to an occluding object, ultrafast imaging aims to leverage the wealth of information usually lost during the capture-time temporal integration. This has revolutionized the fields of imaging and scene understanding, opening up new possibilities, but also discovering new challenges. 

In this paper, we provide an in-depth overview of the most significant works in this domain. We concern ourselves mostly with works in the areas of computer graphics and computer vision; as such, we deal only with visible light and infrared. For other techniques that make use of different wavelengths (such as microwaves, or techniques operating in the terahertz domain), we refer the reader to other excellent sources such as the recent survey by Satat and colleagues~\shortcite{Satat2016}. Similarly, another recent survey~\cite{Bhandari2016} offers an overview of the field from a signal-processing perspective. From our graphics and vision view, we adopt the commonly used term \textit{transient imaging}, referring to  imaging techniques fast enough to capture transient information of light transport, as opposed to traditional techniques that capture steady-state information (such as regular images). 

We have structured our work as follows: First, we introduce \textbf{capture} techniques in \Sec{capture}, separating techniques that directly obtain transient information (such as the previously mentioned femto-photography, or the recent interferometry-based works), from techniques that \textit{reconstruct} that information from a sparse set of measurements, usually sacrificing temporal resolution (such as recent approaches based on time-of-flight (\ToF) cameras). In \Sec{forward_models} we proceed to discuss works whose main goal is to \textbf{analyze} transient light transport, both in the primary and frequency domains. We additionally discuss techniques involving spatio-temporal coding and modulation. In \Sec{applications} we offer a cross section of existing techniques from an \textbf{applications} point of view. Again with a focus on graphics and vision, we subdivide this section in geometry reconstruction, motion estimation, and material estimation; a common problem in most of the applications discussed is the \textit{multipath interference (MPI) problem}, which is tackled from many different angles. With the establishment of transient imaging, the \textbf{simulation} of time-resolved light transport is becoming an increasingly important tool, which we cover in \Sec{simulation}. Last, \Sec{discussion} offers some final conclusions and discussions.

%% file: src20_capture.tex
%!TEX root = src_main.tex
\section{Capture}
\label{sec:capture}

The interaction between light and matter is described as a linear operator by the light transport equation~\cite{Ng2003}:
\begin{equation}
\sImage = \sTransport\sLight,
\label{eq:transport}
\end{equation}
where $\sImage$ is the 2D image (as a column vector of size $\sImageSize$) captured by the camera, $\sLight$ is the vector of size $\sLightSize$ representing the scene illumination, and $\sTransport$ is the scene transport operator encoded as a $\sImageSize \times \sLightSize$ matrix. 
\Eq{transport} assumes that the light transport has reached steady-state. In its transient form~\cite{OToole2014},  incorporating the temporal domain to the light transport equation yields:
\begin{align}
\sTransImage(t) & = \int_{-\infty}^{\infty}\sTransTransport(\tau)\sTransLight(t-\tau)\diff{\tau} \nonumber \\
& = (\sTransTransport \ast \sTransLight)(t), 
\label{eq:transientTransport}
\end{align}
where $\sTransImage(t)$ stores the light arriving at time $t$, $\sTransLight(t)$ is the time-resolved illumination function at instant $t$, and $\sTransTransport(t)$ is the transport matrix describing the light transport with a time-of-flight of exactly $t$. Note that from here on all terms are time-dependent. The second equality represents the convolution in the temporal domain between $\sTransTransport$ and $\sTransLight$. 
In practice, the transient image cannot be captured at instant $t$ directly, given physical limitations of the sensor. Instead, the signal is also convolved by the temporal response of the sensor $\sTransSensor(t)$ centered at $t$ as:
\begin{align}
\sTransImage(t) & = \int_{-\infty}^{\infty} \sTransSensor(t-\tau) (\sTransTransport \ast \sTransLight) (\tau) \diff{\tau} \nonumber \\
& = (\sTransSensor(t) \ast \sTransTransport \ast \sTransLight) (t), 
\label{eq:capturedTransienttransport}
\end{align}

For transient imaging, we are interested in computing the transient image $\sTransImage(t)$ corresponding to the impulse response of the transport matrix $\sTransTransport$ (i.e. \Eq{transientTransport}). This would effectively mean that the illumination $\sTransLight(t)=\delta_0(t)$ and sensor response $\sTransSensor(t)=\delta_t(t)$ are Dirac deltas centered on 0 and $t$ respectively. 
In order to capture this impulse response $\sTransTransport$, several approaches have been presented, depending on the type of illumination and sensor response used. If we focus only on illumination, the main lines of work have used either impulse illumination, or coded illumination. 
In the case of the former, techniques have used either ultrafast imaging systems to directly record light transport (\Sec{impulse}), or phase interferometry to recover the propagation of light (\Sec{interferometry}).
%The former has been generally imaged using ultrafast imaging systems to directly record light transport (\Sec{impulse}), or using phase interferometry to recover the propagation of light (\Sec{interferometry}). 
In the case of the latter, the coded illumination
% The later, on the other hand, 
has been usually correlated with the coded sensor response, allowing to recover the time-resolved response by means of post-capture computation (\Sec{phase}). A comparison of selected capture systems, including their spatio-temporal resolution, is summarized in \Tab{capture}.

\input{src20t_table}
\input{src21_capture_impulse}
\input{src23_capture_interferometry}

\input{src22_capture_phase_tof}

\input{src24_capture_discussion}

%% file: src20t_table.tex
%!TEX root = src_main.tex
\begin{table*}[]
\centering
\begin{tabular}{lllll}
\textbf{Work}           & \textbf{Technology} & \textbf{Spatial Res.} & \textbf{Temporal Res.} & \textbf{Cost}                   \\ \hline
\cite{Gkioulekas2015} & Interferometry      & 655x648               & 33 fs                  & several hours                  \\ 
\cite{Heshmat2014}    & Streak camera       & 56x12                 & 0.3-5.7 ps             & 6 s                           \\
\cite{Velten2013}     & Streak camera       & 672x600               & 0.3-5.7 ps             & 2 h                             \\
\cite{Tadano2015}     & AMCW                & 160x120               & 10 ps                  & linear with temporal resolution \\
\cite{Gao2014}        & Streak camera       & 150x150              & 10/20 ps               & $10^{-11}$ s                    \\
\cite{Laurenzis2014area}  & Laser-gated         & 1360x1024             & 66.7 ps                & ?                               \\ 
\cite{Gariepy2015}    & SPAD                & 32x32                 & 67 ps                  & 240 s                           \\
\cite{Peters2015}     & AMCW                & 160x120               &  $\sim100$ ps                   & 0.05 s (real-time)           \\  \cite{Kadambi2013}    & AMCW                & 160x120               & 100 ps                 & 4 s                             \\
\cite{Busck2004}       & Laser-gated         & 582x752               & 100 ps                 & 32K shots                       \\
\cite{Heide2013}      & AMCW                & 160x120               & 1 ns                 & 90 s (+ hours postprocess)          \\
\cite{Li2012}         & Laser-gated         & 240x240 (*)         & 20 ns                  & 76K shots                       \\
\cite{Laurenzis2007}  & Laser-gated         & 696x520  & 40 ms (*)              & 240 shots                       \\\cite{Dorrington2007} & AMCW                & 512x512               & 0.5 mm (depth)          & 10-20 s                        \\ 
\cite{Hebert1992}     & AMCW          & 256x256 (scan)        & 0.98 cm (depth)         & 0.5 s                           
\end{tabular}
\caption{Selected representative capture setups. Each different technology presents a variety of spatial and temporal resolutions, as well as a range of capturing times. Note that the work by Laurenzis et al.~\protect\shortcite{Laurenzis2007} is targeted for long range depth reconstruction, which imposes a very long exposure time. Note also that Li et al.'s approach~\protect\shortcite{Li2012} is able to reconstruct $240^2$ pixels from a single captured one. }
\label{tab:capture}
\end{table*}

%% file: src21_capture_impulse.tex
%!TEX root = src_main.tex
\subsection{Straight temporal recording}
\label{sec:impulse}

In theory, the most straightforward way to capture the impulse transport matrix $\sTransTransport$ is to use an imaging system with impulse illumination, and extremely short exposure times. However, this is very challenging in practice. First, the signal-to-noise ratio (SNR) is extremely low, since very few photons arrive during the exposure time. On top of this, ultrashort illumination pulses are required, to avoid the effect of the convolution on the transport matrix. Further, ultrafast imaging systems (in the order of nano to picosecond resolution) do not exist for high-resolution, two-dimensional imaging.

%this introduces several challenges~\cite{Velten2013}: \textit{i)} Ultrashort illumination pulses are required, to avoid the convolution of the transport matrix; \textit{ii)} we need ultrafast imaging systems (in the order of nano- o picosecond resolution), which do not exist for high-resolution bidimensional imaging, and \textit{iii)} 

Ultrashort (impulse) illumination is in general achieved by using laser-based illumination, such as femtolasers~\cite{Velten2013}. 
Several different approaches have been proposed to capture transient light transport, and to mitigate the challenge imposed by the extremely low SNR. In the following, we categorize these works according to the imaging system they are based on. Note that we concern ourselves with ultrafast imaging techniques focusing on time-resolved light transport, recording either a single bounce (e.g., for LIDAR applications), or multiple scattering. Other impulse-based ultrafast imaging techniques, e.g., based on pulse stretching~\cite{Nakagawa2014,Lau2016}, are not discussed in this work.

Conceived for range imaging, \emph{laser gated viewing}~\cite{Busck2004} exploits the repeatability of light transport in a static scene by sequentially imaging a set of frames. An ultrashort laser pulse is synchronized with an ultrafast camera equipped with a highly sensitive CCD, which images  photons arriving during very small temporal windows (in the order of a few hundred picoseconds). Each frame is computed independently, by sliding the imaging window. This system was later extended to use area impulse illumination in the context of non-line-of-sight imaging~\cite{Laurenzis2014area}. In order to improve the SNR, hundreds of measurements  are required for each frame. Since each frame needs to be computed independently, gated imaging scales linearly with the number of frames. In order to improve convergence on range imaging (i.e. focusing on single scattering), approaches such as range-gates coding were developed, where the gate response is continuously modulated over a long time window, then reconstructed using intensity analysis~\cite{Laurenzis2007,Zhang2011,Laurenzis2011}, compressed sensing (CS) %~\cite{Donoho2006} 
on the temporal domain by random temporal gating~\cite{Li2012,Tsagkatakis2012,Tsagkatakis2013,Tsagkatakis2015}, or hybrid approaches combining both techniques~\cite{Zhang2012,Dai2013}. This can reduce the number of measurements to just two, for 13-bit range images. An in-depth comparison of these approaches was done by Laurenzis and Woiselle~\shortcite{Laurenzis2014cs}. 
Also relying on CS, Li et al.~\shortcite{Li2012} obtained full 2D transient images using a gated approach with a single pixel detector. 

Systems based on arraying \emph{avalanche photodetectors (APD)}~\cite{Charbon2007}, and in particular \emph{single photon avalanche diodes (SPAD)}~\cite{Kirmani2014,Gariepy2015}, allow reducing significantly both capture times, and the required power of the light source, due to the single-photon sensitivity of the photodetectors. These systems are relatively simple and low-cost, and use eye-safe illumination. On the down side, they yield lower spatial resolutions, and require higher exposure times (in the order of tens of picoseconds). 

\begin{figure}[t]
	\centering
	\includegraphics[width=0.9\columnwidth]{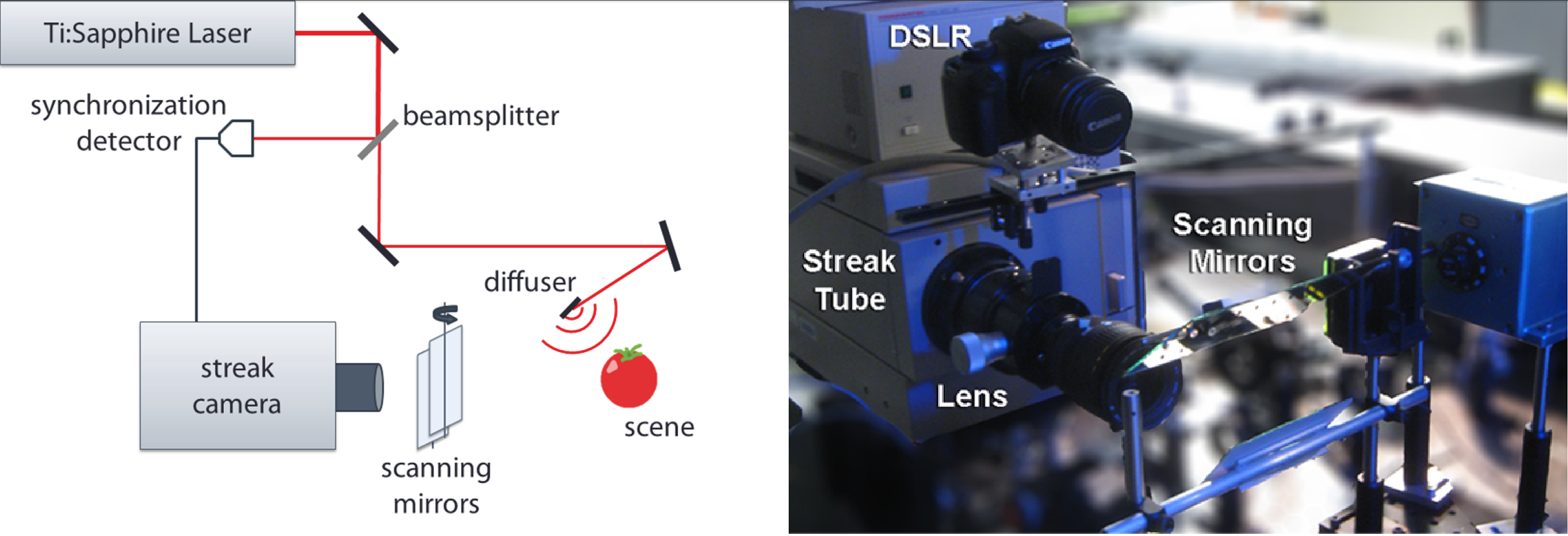}
	\caption{
	\label{fig:femto} \emph{Left:} Diagram of the \emph{femto-photography} setup. A laser, synchronized with the streak camera by means of a beam splitter and a synchronization detector, illuminates the scene after hitting a diffuser; the photons scattered towards the camera are imaged by the streak sensor, which captures a 1D video at picosecond resolution. In order to capture the full 2D time-resolved video, a set of rotating mirrors are used to scan the scene along the y-axis. \emph{Right:} Photography of the setup, the DSLR is used to capture a conventional image of the scene (image from \protect \cite{Velten2013}). }
\end{figure}
In order to improve the temporal resolution, Velten et al.~\shortcite{Velten2012sig,Velten2013,Velten2016} used a streak camera~\cite{Hamamatsu} as imaging device. A streak tube sacrifices one spatial dimension (the y-axis) of the sensor, and uses it to encode the time of arrival of photons. This is done by transforming photons into electrons using a photocathode. The electrons are then deflected at different angles as they pass through a microchannel plate, by means of rapidly changing the voltage between the electrodes. The CCD finally records the horizontal position of each pulse and maps its arrival time to the vertical axis. This effectively records a 1D video of transient light transport, with a temporal resolution of about two picoseconds. In order to record a 2D video, a rotating mirror progressively scans the scene along the vertical axis. \Fig{femto} shows a diagram of the setup, which they called \emph{femto-photography}. As opposed to gated imaging, acquisition times are no longer linear with time resolution, but they scale linearly with the vertical resolution, although it requires repeated captures to get a decent SNR. 
In order to capture the three $x-y-t$ dimensions simultaneously, Heshmat et al.~\shortcite{Heshmat2014} encoded the $x-y$ spatial domain into a single  dimension on the streak sensor, by using a tilted lenslet array. 
Gao et al.~\shortcite{Gao2014}, on the other hand, added a digital micromirror device with a pseudo-random binary pattern encoding the spatial dimension lost in the streak sensor. The transient image is then recovered using sparse reconstruction techniques. 

%% file: src23_capture_interferometry.tex
%!TEX root = src_main.tex
\subsection{Interferometry-based imaging}
\label{sec:interferometry}

\begin{figure}[t]
	\centering
	\includegraphics[width=0.9\columnwidth]{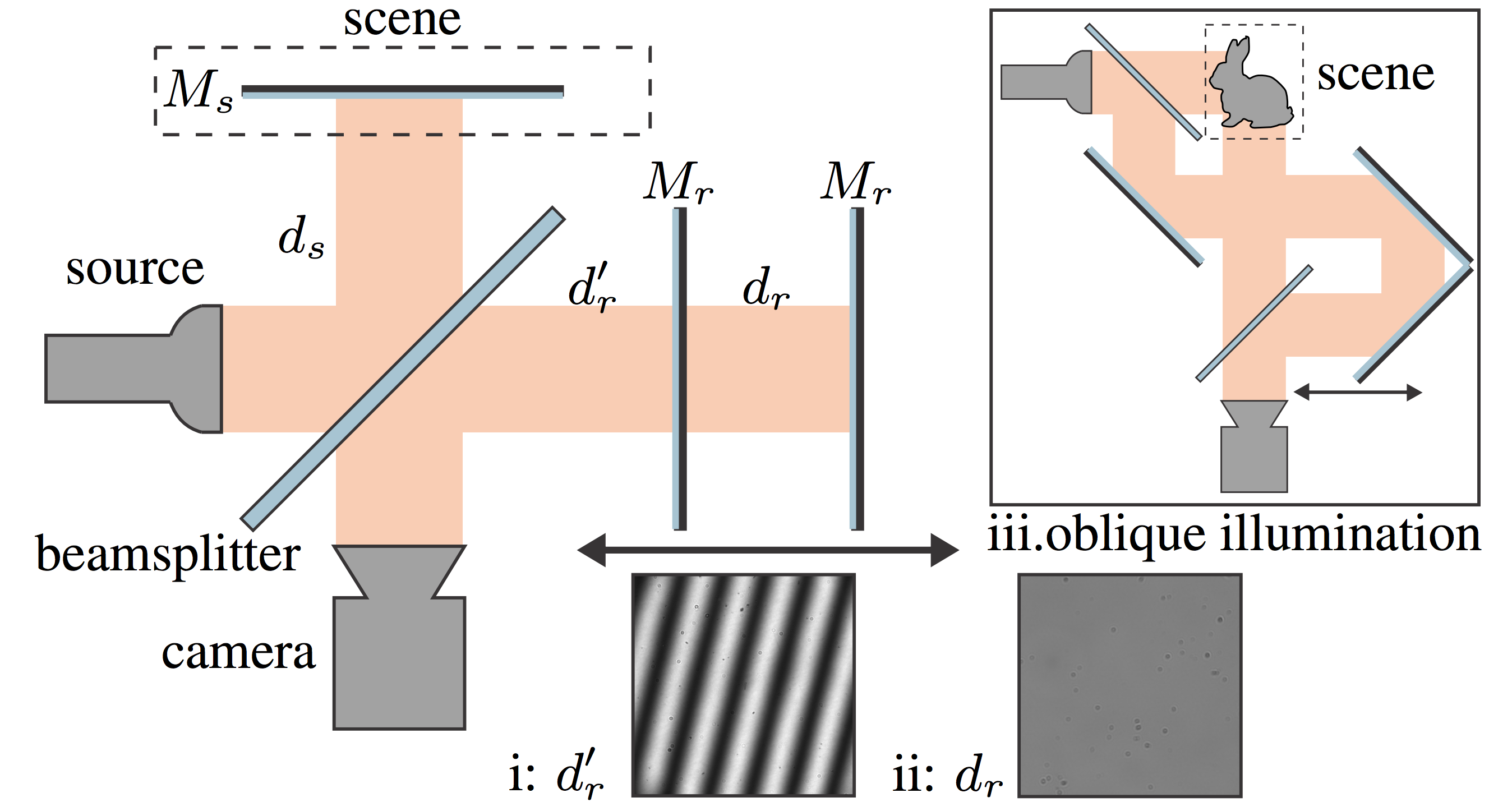}
	\caption{
	\label{fig:interferometry} Michelson interferometer. The beamsplitter sends an input beam to two mirrors $M_r$ and $M_s$ at distances $d_r$ and $d_s$. The split beams reflect back and recombine at the beamsplitter, then are imaged by the camera. Insets $i$ and $ii$ show two results with interference ($d'_r \approx d_s$), and without ($d_r \neq d_s$). Inset $iii$ shows an alternative setup for oblique illumination (image from \protect \cite{Gkioulekas2015}). }
\end{figure}

Interferometry-based imaging techniques rely on creating interference between electromagnetic fields. Not very many works exist in the fields of graphics and vision that take advantage of this methodology. Although its path length resolution is very high (higher than femto-photography by at least an order of magnitude), it presents limitations regarding field of view, depth of field and time of capture. Moreover, it is extremely sensitive to even micron-scale vibrations. Abramson introduced the first light-in-flight visualizations using holography, by illuminating a flat surface and a hologram plate with short pulses of light \shortcite{Abramson1978,Abramson1983}. Recently, Gkioulekas et al.~\shortcite{Gkioulekas2015} presented an optical assembly capable of achieving a resolution of 10 microns ($\sim33.35$ femtoseconds)
%(i.e. ~33.35 femtoseconds)  
by using an interferometry approach inspired in optical coherence tomography (see Figure~\ref{fig:interferometry}. At this scale, the authors can visualize effects such as dispersion or birefringence. Kadambi and colleagues~\shortcite{Kadambi2016b} introduced macroscopic interferometry, relying on frequency sampling of the \ToF data. A key advantage of this novel approach is its implicit resolution of the MPI problem (\Sec{mpi}); instead of having to disentangle phases as in traditional \ToF (which is a hard, non-linear inverse problem, see \Secs{phase}{mpi}), the authors recast the problem as a summation of varying frequencies. In addition, the technique is more robust at low SNR levels. On the downside, the resolution of the system is limited to meter-scale ranges. 

%% file: src22_capture_phase_tof.tex
%!TEX root = src_main.tex
\subsection{Phase Time-of-Flight}
\label{sec:phase}

\begin{figure}[t]
\centering
\includegraphics[width=0.9\columnwidth]{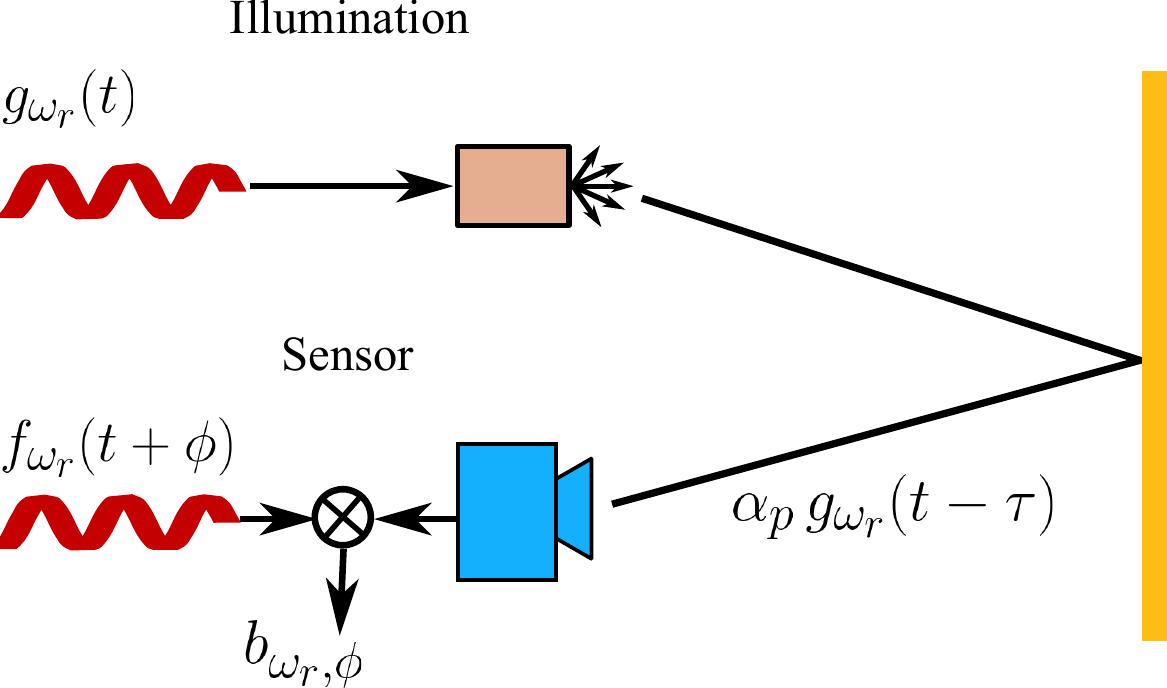}
\caption{\label{fig:AMCWSetup} Basic operation principle of a time-of-flight emitter-sensor setup. Light is amplitude-modulated at the source, constantly emitted towards the scene, and each pixel modulates the impulse response of the observed scene point. By performing cross correlation (integration) of both modulated emitted and received signals, phase differences can be estimated to reconstruct light travel time (image from \protect \cite{Heide2014}).}
\end{figure}

%\begin{figure}[t]
%	\centering
%	\includegraphics[width=0.9\columnwidth]{figures/PMDPrototype}
%	\caption{\label{fig:PMDPrototype} Hardware description of first PMD device \julioc{Placeholder image from \protect \cite{Schwarte1997}, partially %redundant with \Fig{AMCWSetup}, we should mix both in a single image, or remove redundant parts from each one}}
%\end{figure}

Phase-based time-of-flight (P-ToF) imaging, also called correlation-based time-of-flight (C-\ToF) imaging or simply \ToF imaging, cross-correlates emitted modulated light with frequency $\ToFEmitterFreq$, and the impulse response of a pixel $\alpha_p$, modulated and integrated at the sensor with frequency $\ToFSensorFreq$ (see \Fig{AMCWSetup}). In its most typical continuous form (also known as \emph{amplitude modulated continuous wave (AMCW)} systems\footnote{Note that we use the term \emph{AMCW} when referring to these specific sensors, whereas we use \emph{\ToF} for general phase-based time-of-flight sensors.}), the camera computes the cross-correlation as:
\begin{equation}
c(t) =\sTransSensor(t) \ast \sTransLight(t),
\end{equation}
with $\sTransSensor(t)$ the radiance received at the sensor, and $\sTransLight(t)$ the emitted signal. These are in general modeled as:
\begin{align}
\sTransSensor(t) &= \alpha_p \cos(\ToFSensorFreq t + \phi) + \beta, \\
\sTransLight(t) &= \cos(\ToFEmitterFreq t),
\end{align}
where $\phi$ is the phase shift at the sensor, and  $\beta$ the ambient illumination. Capturing a set of different phase shifts $\phi$ allows to retrieve phase differences between the emitted and the received signals. These per-pixel phase differences correspond to light travel time, thus encoding distance (depth), and other possible sources of delay.

Early works demonstrated the applicability and performance limitations of this principle for range imaging in robotic environments \cite{Hebert1992,Adams1996}. Due to hardware characteristics, these approaches were limited to a single range detection per shot, requiring systematic and time-consuming scanning of the scene to obtain a full depth map. 
The first prototype that allowed simultaneous scene capture with modulated array sensors was introduced by Schwarte et al. \shortcite{Schwarte1997}, coined under the denomination of \emph{photonic mixer device (PMD)}. 
Lange and colleagues \shortcite{Lange2000,Lange2001} independently introduced a new type of \ToF devices based on demodulation ``lock-in" pixels, operating on CCD technology with modulation frequencies of a few tens of MHz, and allowing real-time range measurements. These technologies opened new avenues of research on applications and challenges imposed by hardware characteristics. 

An important operational aspect of \ToF setups resides in how the emitter and sensor frequencies are paired. Homodyne configurations use the same frequency at both emitter and sensor ($\ToFSensorFreq = \ToFEmitterFreq$), while heterodyne ones use slightly different frequency pairs. While being more complicated computationally, heterodyne setups have been demonstrated to provide better ranging precision \cite{Conroy2009}, allowing up-to sub-millimeter resolution \cite{Dorrington2007}. Additionally, proper calibration of \ToF cameras was demonstrated to play a significant role when mitigating systematic errors on range estimation \cite{Fuchs2008,Lindner2010}. 

%Heide et al. \shortcite{Heide2013} pioneered the formulation of the first general light transport model for phase-based transient image formation. 
%
%This model is built upon an emitter-sensor setup where a constantly emitted light signal is modulated at the source, propagated through the scene, and reaching an array of pixel sensors. 
%
%At the sensor, each pixel modulates, shifts and integrates the received light during a certain exposure time (see \Fig{}). 

%The basic principle of this model is that travel time of light bouncing in the scene and arriving the sensor is translated to a phase shift in the modulated signal. 
%
Beyond traditional range imaging, Heide and colleagues \shortcite{Heide2013} demonstrated that by correlating a set of sensor measurements with different modulation frequencies and phase shifts, a discrete set of per-pixel light travel times and intensities could be reconstructed through optimization, leading to an inferred transient image of the scene. 
However, the number of frequencies and phases required for this reconstruction is significantly higher than the default set provided by \AMCW devices (a few default frequencies and phases vs. hundreds of them). They work around this issue by substituting the built-in light source, signal generator and phase triggering by external elements. This \AMCW-based setup is much cheaper than femto-photography~\cite{Velten2013}; however, it only reaches nanosecond resolution (compared to picoseconds in femto-photography), the signal is reconstructed as opposed to directly captured, and tweaking the off-the-shelf devices requires a significant amount of skilled work\footnote{http://www.pulsr.info/}.

Successive works aimed to overcome different \ToF devices limitations that affect the viability of subsequent reconstruction methods.  Kadambi et al.~\shortcite{Kadambi2013} reconfigured the emitter modulation with custom-coded illumination, which improved conditioning on the optimization by supporting sparsity constraints. This allowed them to recover per-pixel transient responses using a single frequency, instead of hundreds. Recent work by Peters and colleagues \shortcite{Peters2015} introduced a way to generate robust sinusoidal light signals, which allowed them to obtain up to 18.6 transient responses per second using a closed-form reconstruction method.

\ToF sensor noise, together with limited emitted light intensity due to safety and energy issues, make sensor exposure time and lens aperture the two main factors to achieve an acceptable SNR. To support real-time applications, exposure times must be kept short, so the aperture is usually large to capture as much available light as possible. This introduces a shallow depth of field that blurs scenarios with significant depth changes. Additionally, the low resolution of these sensors (e.g. 200x200 for PMDs) affects the spatial precision of the captured data. 
Godbaz and colleagues \shortcite{Godbaz2010} provided a solution to the shallow depth of field by using coded apertures and explicit range data available in the \ToF camera in order to perform defocus, effectively extending the depth of field. 
Xiao et al.~\shortcite{Xiao2015} leveraged the amplitude and range information provided by the \ToF devices to recover the defocus blur kernel and regularized the optimization in those amplitude and range spaces, allowing for defocus and increased resolution.

Regardless of wide apertures, exposure times need to be much longer than a single modulation period $\ToFSensorPeriod = 1/\ToFSensorFreq$, in order to mitigate sensor noise. This causes a pathological problem known as \emph{phase wrapping}. 
Since light travel time is encoded in the phase shift between emitted and received light, the modulation period $\ToFSensorPeriod$ determines the maximum light path length $\sSpeedOfLight\,\ToFSensorPeriod$ that can be disambiguated, with $\sSpeedOfLight$ the speed of light. 
Any light path starting at the emitter that takes longer than this distance to reach a pixel in the sensor will \emph{phase-wrap} $\ToFSensorPeriod$, falling into the same phase shift than shorter paths within subsequent modulation periods. These phase-wrapped light paths produce interference in the measured data, leading to errors in the reconstruction. 
A straightforward way to solve this is to lower the modulation frequency, thus increasing the maximum unambiguous path length. However, this decreases the accuracy obtained for the reconstructed path lengths, leading to less precise depth measurements. Jongenelen et al. \shortcite{Jongenelen2010} demonstrated how to extend unambiguous maximum range while mitigating precision degradation, by exploring different dual combinations of simultaneous high and low modulation frequencies. Recently, the work by Gupta and colleagues \shortcite{Gupta2014} generalized the use of multiple high frequencies sequentially for this purpose in what they denominate micro-\ToF imaging. 
Phase-wrapping is closely related to the widely-studied problem of MPI, where light from multiple light paths is integrated in the sensor resulting in signal interference and thus reconstruction errors. % (see \Sec{mpi} for a detailed explanation) \diegoc{check. Is this the first time we mention it? Otherwise, don't explain it}. Multi-path interference refers to the estimation errors caused by indirect (multi-path) light reaching a sensor pixel at the same time as direct (single-path) light. 
However, this is related to how some physical phenomena (e.g. interreflections, scattering) affect certain applications ---actually affecting other capture methods too---, rather than to operational limitations of the \ToF devices themselves. Please refer to \Sec{mpi} for a more detailed discussion.
%
%\diegoc{Alternative paragraph: Phase-wrapping is closely related to the problem of multi-path interference (MPI), a physical phenomena not related to  operational limitations of the \ToF devices themselves. Please refer to \Sec{applications} for a more detailed discussion.}

Recent works explore novel hardware modifications: Tadano and colleagues \shortcite{Tadano2015} increased temporal resolution beyond the limit of current \ToF devices (around 100 picoseconds), by using arrays of LED emitters spatially separated by 3mm. This effectively corresponds to time shifts of 10 picoseconds. Shrestha and colleagues~\shortcite{Shrestha2016} explored imaging applications synchronizing up-to three multi-view \ToF cameras. To achieve this, they addressed interference problems between the light sources of the cameras, showing how they can be mitigated by using different sinusoidal frequencies for each sensor/light pair. The authors demonstrated applications such as improved range imaging for dynamic scenes by measuring phase images in parallel with two cameras, doubling single-camera frame rate, and mitigating motion artifacts.

%% file: src24_capture_discussion.tex
%!TEX root = src_main.tex
\subsection{Discussion}
\label{sec:captureDiscussion}

In general, transient imaging systems present a trade-off between cost, ease of use, acquisition time, and quality (spatial and temporal resolution), as summarized in \Tab{capture}. 
The impulse-based techniques described in \Sec{impulse}, while equivalent in terms of the main imaging principle, exhibit several differences regarding the mentioned trade-off. Avalanche photodetectors allow for cheap and fast imaging of single-scattered photons, suitable for LIDAR applications; the time resolution achieved is in the order of tens of picoseconds. 
Gated systems, on the other hand, are more costly in terms of measurements, since they image each frame independently, although this cost can be alleviated when focusing on range sensing. In addition, the temporal resolution of these systems ranges from hundreds of picoseconds to milliseconds, allowing for a large variety of application domains.
The systems based on streak cameras offer the highest temporal resolution, in the order of hundreds of femtoseconds, at the expense of longer exposure times to achieve a workable SNR. 
Last, systems based on interferometry (\Sec{interferometry}) yield the highest temporal resolution, but present many shortcomings including motion sensibility, extremely shallow depth of field, and inherent complexity, which make them hard to use in general scenarios.

On the other hand, phase-based time-of-flight systems (\Sec{phase}) provide a more affordable hardware alternative, with shorter acquisition times. 
However, these devices (e.g. PMD, \ToF Kinect) cannot be used off-the-shelf, requiring many modifications in terms of hardware components, electronics, signal generation, or even multi-device configurations, to achieve acceptable results.  Despite considerable hacking, they yield a much lower temporal resolution---usually in the range of nanoseconds---compared to the pico- or femtoseconds obtained with impulse-based and interferometry-based approaches, respectively. Moreover, these systems do not directly acquire temporal information; instead they require post-capture computations to reconstruct the signal, which may incur in errors.

%off-the-shelf devices (e.g. PMD, Time-of-Flight Kinect) are quite limited in terms of internal characteristics, and in general device software does not grant full access to all operational modes, usually being directed towards range imaging applications. Default built-in frequencies and electronic characteristics support much lower temporal resolutions ---nanoseconds vs. picoseconds or femtoseconds obtained with impulse-based and interferometry approaches respectively---, and while array sensors simultaneously acquire full-field information, the spatial resolution is usually low and very prone to noise. In consequence, these devices require many modifications in terms of hardware components, electronics, signal generation, or even multi-device configurations to achieve acceptable results. 
%
%\adrianc{Well, I'm not fully agree with this last problem in comparison of AMCW vs other methods: one would say the same for impulse-based systems, since there are hardly no off-the-shelf interferometry or femtophoto systems. The main prob with phase-systems is rez, and probably that they don't directly acquire the temporal domain, but they need post-capture computation prone to errors.}
%%

%% file: src30_forward_models.tex
%!TEX root = src_main.tex

\section{Analysis of transient light transport}
\label{sec:forward_models}

The data captured by transient imaging devices is affected by the MPI problem: light from different paths arrives at the same pixel in the sensor (see \Sec{mpi}). This makes analysis difficult, and limits the range of applications for transient imaging. 
%
%The situation is even worse with correlation-based sensors (\Sec{phase}) \adrianc{We need to choose how are we calling these sensors!}, which in its simplest configuration are just able to deliver range; unfortunately, the range estimate is in general very sensible to errors, due to the problem known as \emph{multipath inteference} (MPI), where light from different paths arrive to the same pixel in the sensor (we explain the problem in detail in \Sec{geometry}). 
%
%To overcome this, it is needed to process the captured data before feeding the different applications, decomposing the different illumination components so that the information in e.g. indirect illumination affects the direct illumination. For that, different approaches have been proposed, which can be categorized in two main directions: the first one seeks to exploit the sparsity (or compressibility) of the light transport data in the temporal domain; the second, on the other hand, analyzes the transient light transport on a Fourier-basis. In addition to these two main approaches, other works exploit the separability of light transport by spatio-temporal coding during capture. 

To overcome this, different approaches have been proposed, which can be categorized in two main directions: the first one seeks to exploit the sparsity (or compressibility) of the light transport data in the temporal domain; the second analyzes transient light transport in the Fourier domain. In addition, other works exploit the separability of light transport by spatio-temporal coding during capture. 

\subsection{Sparsity and compressibility}
\label{sec:sparsity}
%
%Talk about the sparsity assumption, in both gradient [Wu,Heide] and primal domains [the rest]. Talk about the phenomenological models using both assumptions: k-sparse (and 2-sparse), delta + exponential, Gaussians...
A common approach for transient light transport analysis leverages the sparsity of light transport, either in the temporal domain or its derivatives, or by modeling it as a \emph{compressive signal} in some alternative basis, according to the particular scene and the targeted application. Depending on these forms of sparsity, a number of phenomenological models have been developed, from explicitly modeling physically-based priors (e.g. $K$-sparse models in the primal domain), to other models based on observation of the captured phenomena. 

\begin{figure*}[t]
	\centering
	\includegraphics[width=0.9\textwidth]{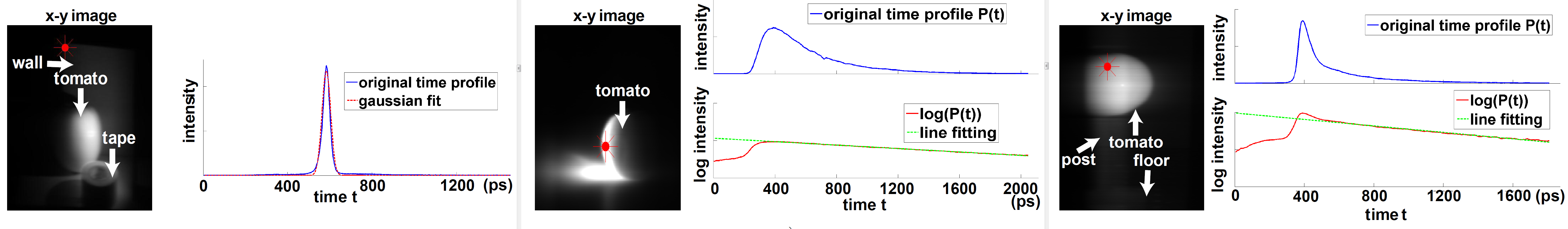}
	\caption{ 
	\label{fig:decomposition} Decomposition on different components of transient light transport. \emph{Left:} A Gaussian profile closely approximates the convolution of the sensor's temporal PSF and the direct illumination impulse response. \emph{Middle:} For a point where subsurface scattering is dominant, the time profile decays exponentially over time. \emph{Right:} A time profile showing both direct and subsurface light transport, which results in the sum of a Gaussian impulse (direct illumination) and an exponential decay (SSS) (image from \protect \cite{Wu2014}). }
\end{figure*}
\paragraph{K-sparse responses model}
One of the most common approaches involves considering the time-profile as a mixture of $K$ Dirac $\delta$-functions in the temporal domain, where each light interaction with surfaces is represented as an impulse response:
\begin{equation}
\sTransTransportPixel(t) = \sum_{i=1}^{K} \alpha_i \delta_{t_i}(t),
\label{eq:kimpulse}
\end{equation}
where $\sTransTransportPixel(t)$ is the time-resolved transport response at a single pixel, $\alpha_i$ is the intensity reaching the sensor, and $\delta_{t_i}$ is the impulse response at instant $t_i$. 

This model forms the basis of gated-laser (\Sec{impulse}) and correlation-based (\Sec{phase}) range methods, which assume a single delta response ($K=1$) on the first scattering event with the nearest surface. This single-bounce assumption is however far from robust, since it assumes scattering on opaque surfaces with no indirect illumination, and thus suffers from the MPI problem. To partially alleviate this, other approaches include additional sparse responses for other interreflections, including an arbitrary large number of responses ($K>1$)~\cite{Kadambi2013,Kirmani2013,Bhandari2015,Qiao2015,Kadambi2015,Peters2015}. For the particular case of $K=2$, this approach allows for a very fast reconstruction ~\cite{Dorrington2011,Godbaz2012,Adam2015,Naik2015}. %\adrianc{Somewhere we need to note that most $K$-sparse models don't capture the integration of delta signals (some do, such as gated sensors), but this delta function convolved with a sinusoid wave. Moreover, at some point it'd be good to state that these works, in general, model these signals as a phasor, which allows for compact representation. Gupta later uses this property to model all transient light transport as phasor operations... I'd say to include the phasor stuff in Sec. 2.2, since the 4-phases approximation for range imaging uses this approach. }
 
Wu et al.~\shortcite{Wu2012cvpr,Wu2014} noted that in streak images this $K$-sparse model is actually convolved by the sensor temporal point spread function (PSF) (\Fig{decomposition}, left), which presents a Gaussian shape~\cite{Velten2013}. Following this observation, the authors model \Eq{kimpulse} as a sum of Gaussians instead:
\begin{equation}
\sTransImage(t) = \sum_{i=1}^{K} \alpha_i G_{t_i}(t,\sigma)
\label{eq:gaussian_impulse}
\end{equation}
where $G_{t_i}$ is the Gaussian centered at $t_i$, with standard deviation $\sigma$ dependent on the imaging system. Note that \Eq{gaussian_impulse} does not model the transport operator $\sTransTransportPixel(t)$, but the imaged pixel at the transient image $\sTransImage(t)$. Using Gaussians instead of delta functions breaks the (primal) sparsity assumption; however, as noted by the authors, the signal is still sparse in the gradient domain for direct transport, as well as specular reflections and transmissions. This allowed them to separate direct and global transport components (interreflections plus subsurface scattering). Further work used this model to reconstruct the full transient image based on correlation-based sensors~\cite{Heide2013}, while Hu et al.~\shortcite{Hu2014} improved Wu's technique by using a more robust method based on convolutional sparse coding. 

The models discussed so far assume impulse light transport, where all scattering events occur between perfectly opaque or transmissive surfaces and light tavels freely through empty space. Since indirect illumination is a continuous signal, it would be impractical to model it using the $K$-impulse model (or its Gaussians-based version). Two other models, discussed below, address subsurface (volumetric) scattering, as well as diffuse interreflections.

%\adrianc{Should we give details on how these models are computed? In general they are some sparse regularization, but e.g. Peters include some nice other features (Piserenko estimate for the k-delta approach).}

%The main drawback of this approximation is that the sparse impulse approximation only matches the signal of direct transport, as well as specular reflections and transmissions. For diffuse light transport (Lambertian indirect transport, as well as subsurface or volumetric scattering), the sparsity assumption on the primal domain is broken, and therefore the model is not robust to these types of transport components. \diegoc{as written, this clashes with the previous paragraph}

\paragraph{Exponential volumetric models}
In the presence of translucent objects or participating media, indirect illumination plays an important role. 
%The presence o in general breaks this assumption, and the $K$-sparse models fail to describe these effects. 
%
%For subsurface scattering where the Gaussians model does not hold, Wu and colleagues used instead a decaying exponential function. 
For the case of translucent, optically thick objects, Wu et al.~\shortcite{Wu2012cvpr,Wu2014} empirically observed that a single exponential decay accurately models subsurface scattering as:
\begin{equation}
\sSSS(t) = \exp(\log(L_i) + \gamma t),
\end{equation}
where $L_i$ is the incident illumination and $\gamma$ is the scattering coefficient of the object (\Fig{decomposition}, middle and right). 

For light transport in more general participating media (i.e. relaxing the assumption of optically thick media), Heide et al.~\shortcite{Heide2014} used a similar exponential decay assumption; however, since most sensors have a Gaussian PSF, they approximated this decay by using an exponentially modified Gaussian. This allowed them to model transient light transport in participating media as a mixture model of these type of Gaussians, which provide a compressive base for transient light transport in such media. 
Interestingly, a similar set of time-resolved exponentially modified Gaussians, integrated in the temporal domain, lies at the core of the quantized diffusion model~\cite{DEon2011} for rendering high-quality subsurface scattering.

\paragraph{Exponential-based diffuse illumination}
Freedman et al.~\shortcite{Freedman2014} observed that the Lambertian indirect temporal response $\sIndirectDiffuse(t)$ also presents a smooth exponential shape, although different from Wu's. They modeled it as:
\begin{equation}
\sIndirectDiffuse(t) = A t^{\alpha} e^{−\beta{t}},
\end{equation}
where $A$, $\alpha$ and $\beta$ depend on the geometry and reflectance of the underlying scene, and $t\geq0$. For $t<0$, we have $\sIndirectDiffuse(t)=0$.

With this approach, the authors modeled transient light transport as a sum of the impulse-like transport ($K$-sparse model) for direct and specular paths, plus the Lambertian indirect illumination as:
\begin{equation}
\sTransTransportPixel(t) = \sum_{i=1}^{K} \alpha_i \delta_{t_i}(t,\sigma) + \sum_{j=i}^{K_L} \sIndirectDiffuse_j(t-t_j),
\end{equation}
with $t_j$ the shortest light path of the diffuser.  Unfortunately, in this formulation transient light transport is no longer sparse neither in the primal nor in the gradient domains. However, the impulse functions and the exponential decays form again a compressive base.

%\adrianc{FreedmanECCV2014 - 3 BhandariOptLett2014 - 77 == Delta responses + Exponential decay for diffuse transport.}

%Based on the $K$-sparse responses models, Wu et al.~\cite{} realized that for most imaging systems, the impulse responses are convolved by the temporal PSF of the sensor. This breaks the assumption of sparsity on the primal domain, although the signal is still compressive, i.e. sparse in a certain base, in this case a Gaussian mixture. 

%Wu et al.~\cite{Wu2014} modeled the temporal profile and an exponentially decaying function for subsurface scattering. They also presented an algorithm to separate the profile at each pixel into these three components. The work of Hu et al.~\cite{Hu2014} builds on a similar model of the temporal profile.

%Heide et al.\cite{Heide2014} representing the transient signal of a pixel as a mixture of exponentially modified Gaussians. Included not only direct transport, but also within participating media. Their goal is to address MPI in the presence of media. \adrianc{Check more...} 
%
%\adrianc{We should relate Heide's Guassian mixture with d'Eon's~\shortcite{DEon2011} quantized diffusion, which in the end is a sum of time-resolved Gaussians.}

%
\begin{figure}
\centering
\includegraphics[width=\columnwidth]{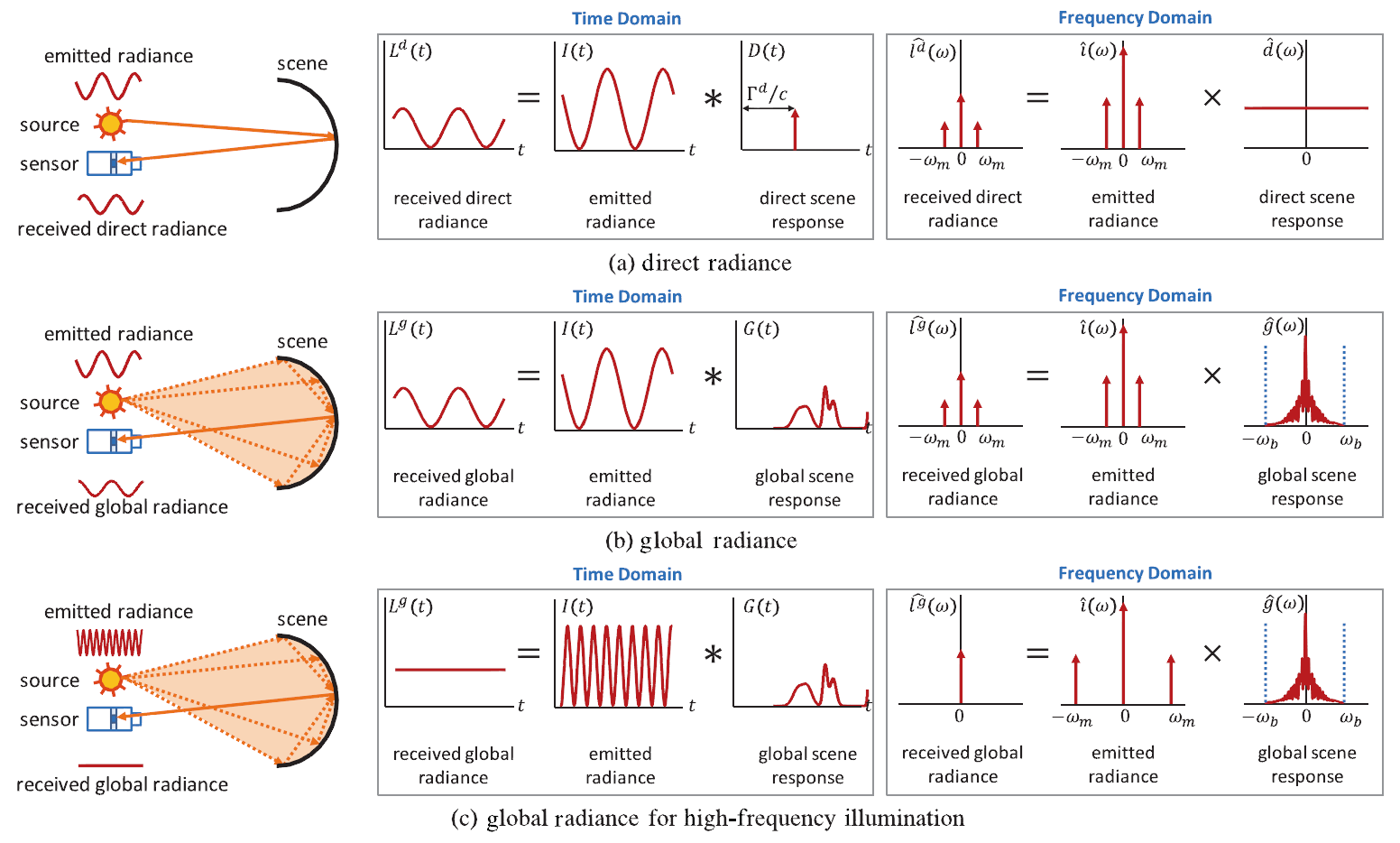}
\caption{\label{fig:phasor_frequency} Frequency analysis of light transport using AMCW systems, modeled using a phasor representation of the sinusoidal signal. In the primal domain, the response is obtained by convolving the emitted radiance with the response of the scene, which is a delta response for direct transport (a), and might get arbitrarily complex in the presence of multiple interreflections (b,c). This response is in general smooth, and therefore bandlimited in the frequency domain. This means that for high-frequency modulated emitted radiance, the global response contains only the DC component (image adapted from \protect \cite{Gupta2014}).}
\end{figure}

\subsection{Frequency domain}
\label{sec:fourier}
Fourier-based analyses and techniques have been developed to gain fundamental insights about the information encoded in the temporal domain, and as a tool for effective capture and processing of transient data. 
Wu et al.~\shortcite{Wu2012eccv} extended the frequency analysis on the incoming steady-state light field~\cite{Durand2005} by analyzing the full 5D time-resolved light field. Their analysis revealed a cross-dimensional information transfer between domains. They further demonstrated potential applications of this analysis by prototyping a transient-based bare sensor imaging system.
Lin and colleagues~\shortcite{Lin2014,Lin2016} observed that capturing transient light transport using AMCW sensors with a homodyne setup is equivalent to sampling transient light transport in the frequency domain. Based on this observation, they recovered the temporal radiance  profile by means of the inverse Fourier transform of the captured data. 
Kadambi et al.~\shortcite{Kadambi2015} showed that the formulation used in AMCW (based on the correlation of sinusoid waves, encoded as phasors), and frequency-domain Optical Coherence Tomography (OCT)~\cite{Huang1991} are analogous. This allowed them to use the well-studied methods in OCT to capture transient light transport. One of the advantages of these techniques is that they do not require sampling in phase; as the authors demonstrate, this allows to use any sensor (including traditional CMOS-based cameras), and might eventually lead to sub-picosecond resolution captures with standard, low-cost time-of-flight sensors. 

Finally, Gupta et al.~\shortcite{Gupta2014} used frequency analysis on top of their phasor-based model of light transport. The authors demonstrated that for high-frequency correlation-based sensing, diffuse interreflections vanish due to the band-limited nature of such interreflections. This allowed them to avoid the multipath interference problem by capturing direct light transport with high-frequency sparse impulses, while lower modulation frequencies were used to capture indirect transport (\Fig{phasor_frequency}).

\subsection{Spatially-coded capture}
A third  approach involves spatial light coding, coupled with temporal modulation during capture. These strategies allow to analyze (decompose) light transport in scenes, without explicitly knowing its temporal profile.
Naik et al.~\shortcite{Naik2015} exploited the relationship between direct and indirect illumination when using high-frequency spatial illumination patterns. They leveraged the work of Nayar et al.~\shortcite{Nayar2006} to decouple both components, significantly improving range imaging. This work however needs an external projector to perform the separation. To overcome this problem, Whyte et al.~\shortcite{Whyte2015} derived a theoretical framework to perform spatially-modulated separation using phases and amplitudes, allowing their use in AMCW systems. They additionally obtained an optimal set of projection spatial patterns. Also based on spatial modulation, O'Toole and colleagues~\shortcite{OToole2014} used optical probing of light transport~\cite{OToole2012}, which allows not only direct-indirect separation, but also decomposing high- and low-frequencies of indirect transport. %\diegoc{no habia un paper de Ravi en una de vision, casi simultaneo, donde hacia algo muy similar a OToole?}\adrianc{Si, un ECCV, pero no usaba exactamente lo que O'Toole [2012], y [OToole14] si usa esa tecnica en concreto.}

\subsection{Discussion}
Analyzing transient light transport is a key step towards producing practical applications. Exploiting sparsity in the primal domain has been a common, flexible approach. However, although the $K$-sparse models have physical meaning, they become impractical for diffuse interreflections, including volumetric and subsurface scattering. 
To overcome this, phenomenological models based on empirical observations have been developed, and shown to provide a good base for compressive decomposition. 
On the other hand, moving from the primal to the Fourier domain helps extend the range of potential analyses, allowing more principled decompositions, or using tools from well-established fields (e.g. inverse Fourier transform, or frequency-based OCT). This helps reduce the need for heuristic bases to represent light transport. 
Interestingly, some other works follow a different approach, making use of spatial light modulation together with temporal modulation, leading to very robust light transport decomposition at the cost of additional effort and machinery during capture.

%% file: src40_applications.tex
%!TEX root = src_main.tex
\section{Applications}
\label{sec:applications}

The temporal resolution offered by transient data has opened up a wealth of possibilities in terms of applications. Long-standing vision and graphics problems, such as depth recovery in the presence of interreflections, or light component separation, have received renewed attention in light of this new data~\cite{Raskar2008}. Here we delve into the areas which have benefited most from transient imaging in the realm of graphics and vision, but applications extend to fields such as medical imaging, surveillance, or atmospheric sciences, to name a few.

\input{src41_geometry_reconstruction}

\input{src42_motion_estimation}
\input{src43_material_estimation}

%\input{src34_illumination_decomposition}
%\input{src35_bare_sensor_imaging}

%% file: src41_geometry_reconstruction.tex
%!TEX root = src_main.tex
\subsection{Geometry reconstruction}
\label{sec:geometry}

The most prevalent application of transient imaging is the recovery of depth information from the scene. More recently, ultrafast transient data has opened the door to recovering not only depth but full geometry information, i.e., including non-line-of-sight areas (due to occlusions or the presence of a dense medium). This requires being able to drop the assumption that light only bounces once before reaching the camera; effectively, this means having the ability to separate the paths followed by the different light rays, that is, solving the  multipath interference problem. While a large body of work on traditional range imaging via \ToF exists~\cite{Hansard2012,Remondino2013}, we place the focus here on the recent approaches exploiting ultrafast data. The reader may also refer to specialized tutorials on \ToF imaging~\cite{Kolb2010,Kadambi2015CursoICCV}.
% The problem of geometry estimation is closely related to that dealt with in the previous subsection, and as such many techniques are shared. In the area of geometry reconstruction, transient data has been traditionally used for range imaging. However, recently, ultrafast transient data \belenc{I wrote ultrafast but careful with this. We may want to avoid. Actually, will we have a definition of what transient data is? :) We should.} has opened the door not only to ``around the corner" reconstructions, but also to finding a solution to long-standing problems such as the presence of multiple interreflections (the multi-path interference or MPI problem) and, as a particularly challenging subset of this, imaging through dense (scattering) media. 

\subsubsection{Range imaging} 
\label{sec:range}
Range imaging, that is, obtaining depth information from a scene, has been traditionally achieved via two different methods: those based on multiple viewpoints of the scene\footnote{Structured light approaches, relying on one camera and one projector, can be considered a subset of these~\cite{Gupta2012}.} and obtaining correspondences between them, and those based on time of flight. The principle behind \ToF sensors has been explained above (please refer to \Sec{phase} for a detailed description), and the computation of depth information (i.e. distance to the camera, $d$) from that temporal data is as conceptually as simple as applying $d = vt$, where $t$ is the time light has taken to travel the distance $d$, given its velocity in the medium $v=c \eta$ (where $c$ is the speed of light in a vacuum). More specifically, in the commonly used AMCW cameras, what is measured is the phase shift $\theta$ in the modulation envelope (as well as the amplitude), from which the distance can be obtained as~\cite{Dorrington2011,Kadambi2015}: 
\begin{equation}
d = \frac{v\theta}{4\pi \ToFSensorFreq}, 
\end{equation}
where $\ToFSensorFreq$ is the frequency of the modulation signal. 

\begin{figure*}[t]
	\centering
	\includegraphics[width=\textwidth]{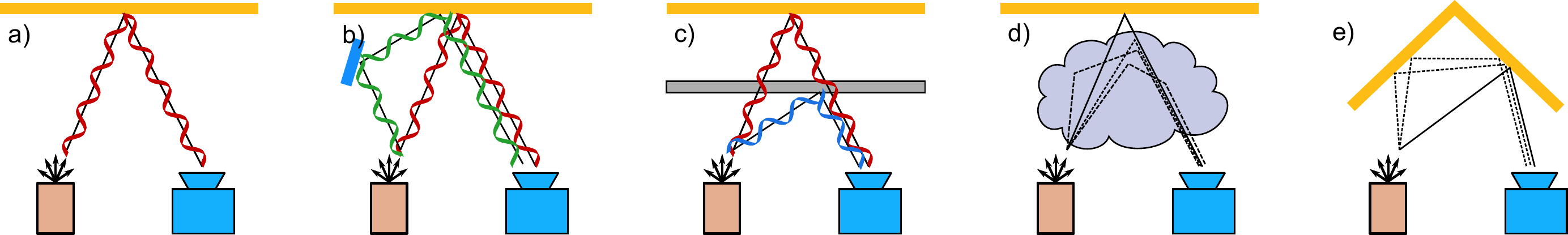} % Alt.: Fig_MPI_Explanation_Placeholder.png from Dorrington et al. 2011
	\caption{\label{fig:mpi} a) Range imaging using a \ToF sensor, where the phase delay of the emitted modulated radiance encodes light time of flight. This formulation works for direct reflection, but presents problems when light from multiple paths is integrated in the sensor, as happens in the presence of specular reflections (b) or tranmission (c). This problem gets even more complicated in the presence of diffuse transport, due to scattering in media (d) or Lambertian diffuse interreflections (e). Image after \protect\cite{Bhandari2014}.}
\end{figure*}

Modern Kinect sensors, for instance, employ this technology to obtain depth information~\cite{Bamji2015}. \ToF cameras allow obtaining relatively low resolution depth images of the scene. Still, they offer a great advantage over previous scan-based (LIDAR) techniques---which measured the time of flight of a laser beam to a point---: the capture speed is in the order of hundreds of frames per second. The main stumbling block of traditional \ToF techniques for range imaging is the underlying assumption that the light only bounces once before reaching the camera. While this may be true in some scenes, in many other cases, and particularly in the presence of concavities (such as the corners of a room), purely specular objects, and scattering media (fog, tissue, etc.), this no longer holds and conventional depth recovery methods fail (see Figure~\ref{fig:mpi}). We cover techniques developed to address this problem in \Sec{mpi}. 
%[Something on dates and commercial systems? XXX]
% Bamji2015: http://ieeexplore.ieee.org/stamp/stamp.jsp?arnumber=6964815 - A 0.13 μm CMOS System-on-Chip for a 512 × 424 Time-of-Flight Image Sensor With Multi-Frequency Photo-Demodulation up to 130 MHz and 2 GS/s ADC

Other means of obtaining transient data, such as direct imagers, have also been applied to the recovery of depth information. Again, if the assumption for each point is that light arriving to the camera has bounced only once, reconstructing depth is trivial, but this is often not the case. Thus, the focus of existing works in the area of direct imagers has been to reconstruct occluded geometry. We cover these works in \Sec{nlos}.

%Finally, even though most current \ToF systems are \emph{phase-based}---i.e., the recorded returning signal measures, using time correlation, a phase-shift which is depth dependent---, Kadambi et al.~\shortcite{Kadambi2015} presented a technique to obtain depth relying only on the frequency of the received signal, a concept analogous to that of frequency-domain OCT.

\input{src41t_table}
\subsubsection{The multipath interference (MPI) problem} 
\label{sec:mpi}
The MPI problem is common for most transient imaging devices, specially in those with long exposure times: for example, in the context of gated-based LIDAR systems (\Sec{impulse}), where a modulated sensor response~\cite{Laurenzis2007,Laurenzis2014cs} is used to robustly acquire depth. However, is in \ToF cameras where the problem is more noticeable.  Some early approaches to solving the MPI problem in ToF cameras targeted in-camera light scattering~\cite{Kavli2008,Schaefer2014}; others targeted also indirect illumination but require placing tags in the scene~\cite{Falie2009}, or made severe assumptions on scene characteristics~\cite{Jamtsho2010}. For an in-depth discussion about the MPI problem from a signal processing perspective, we refer the reader to a recent article by Bhandari and Raskar~\shortcite{Bhandari2016}. A comparative summary of the techniques discussed in this section can be found in \Tab{mpi}.

The work of Fuchs~\shortcite{Fuchs2010} provided a model of MPI for the case in which all distracting surfaces are Lambertian, based on explicitly computing indirect illumination on the estimated depth map and iteratively correcting it. 
% Kavli: http://spie.org/Publications/Proceedings/Paper/10.1117/12.791019
% Schaefer: http://www.franklenzen.de/pdf/schaefer_oe2014_reprint.pdf
% Jamtsho: http://www.isprs.org/proceedings/XXXVIII/part5/papers/67.pdf
% Falie: https://arxiv.org/ftp/arxiv/papers/0909/0909.5656.pdf
Follow-up works aimed at a more general solution targeting the source of the problem: the separation of the individual components when multiple returns are present~\cite{Godbaz2008,Godbaz2009}, also called Mixed Pixel Restoration. These techniques, however, cannot be used with off-the-shelf cameras, since they require measuring multiple phase steps per range measurement (as opposed to the usual four).
% consider citing only one in Godbaz 08 and 09 - check if same
% Godbaz 2008: Mixed pixel return separation for a full-field ranger”, Proc. Proc. Of the Image and Vision Computing New Zealand Conference (IVCNZ’08), Christchurch, New Zealand, ISBN 978-1-4244-3780-1, 1-6 (2008).
% Godbaz 2009: Godbaz, J. P., Cree, M. J. and Dorrington A. A., “Multiple return separation for a full-field ranger via continuous waveform modeling”, Proc. SPIE 7251, 72510T (2009).
Of large relevance is the work of Dorrington et al.~\shortcite{Dorrington2011}, in which the authors proposed a numerical solution that can be employed in off-the-shelf ToF cameras. Shortly after, Godbaz et al.~\shortcite{Godbaz2012} proposed two closed-form solutions to the problem. These two works assume, however, that there are two return components per pixel, and work with two or up to four modulation frequencies. This two-component, dual-frequency approach was generalized by Bhandari et al.~\shortcite{Bhandari2014}. Kirmani et al.~\shortcite{Kirmani2013} targeted simultaneously phase unwrapping and multipath interference cancellation, using a higher number of frequencies (five or more), but at a lower computational cost than previous approaches, thanks to a closed form solution. Still, they assumed sparsity in the recovered signal, and again restricted their model to two-bounce situations ($K=2$, see \Sec{sparsity}). 
% Dorrington: https://core.ac.uk/download/pdf/29198999.pdf
% Godbaz2012: https://ai2-s2-pdfs.s3.amazonaws.com/1c76/587c94e73f4ae2e6b780258ef50b61121226.pdf
% Kirmani 2013: https://www.microsoft.com/en-us/research/publication/spumic-simultaneous-phase-unwrapping-and-multipath-interference-cancellation-in-time-of-flight-cameras-using-spectal-methods/
% Bhandari2014: https://arxiv.org/pdf/1404.1116.pdf

The use of multiple modulation frequencies was also leveraged by Heide and colleagues~\shortcite{Heide2013}. In their case, they used hundreds of modulation frequencies, and proposed a model that includes global illumination. Freedman et al.~\shortcite{Freedman2014} also required multiple frequencies, and proposed a model (not limited to two bounces) which assumes compressibility of the time profile; they solved the problem iteratively via $\mathcal{L}_1$ optimization. Kadambi et al.~\shortcite{Kadambi2013} reduced the number of frequencies required to recover a time profile (and thus depth information) to one, by using custom codes in the emission in combination with sparse deconvolution techniques, to recover the time profiles as a sparse set of Dirac deltas. This technique allowed to recover depth in the presence of interreflections, including transparent objects (Figure~\ref{fig:unicorn}). 

All these works assumed a $K$-sparse transport model (\Sec{sparsity}). It is worth noting, however, that in the case of scattering media being present, a sparse formulation of the time profile is no longer possible. The problem of scattering media is treated in \Sec{scattering}. A slightly different approach was taken by Jim\'enez et al.~\shortcite{Jimenez2014}, who proposed an optimization framework to minimize the difference between the measured depth, and the depth obtained by their radiometric model.
%optimized for the depth correction with which the radiometric model generates the closest depth map to the one measured; 
Convergence to a global minimum was not guaranteed, but a number of examples including real scenes were shown. Hardware modifications are not required.

A different means of eliminating or separating global light transport in a scene was presented by O'Toole et al.~\shortcite{OToole2014}, who made the key observation that transient light transport is separable in the temporal frequency domain (see \Sec{fourier}). This allowed them to acquire and process only the direct time-of-flight component, by using a projector with light modulated in space and time (note that they do not use correlation-based ToF). Gupta et al.~\shortcite{Gupta2014} built on this idea, and proposed a framework termed \emph{phasor imaging}. A key observation is that global effects vanish for frequencies higher than a certain, scene-dependent, threshold; this allowed the authors to recover depth in the presence of MPI, as well as to perform direct/global separation, using correlation-based time-of-flight sensors. Neither Gupta et al.'s work, nor O'Toole et al.'s, imposed the restriction of sparsity of the multipath profile. Neither did Naik et al.~\shortcite{Naik2015}, who also attempted direct/global separation to obtain correct depth in the presence of MPI. A similar approach was followed by Whyte et al.~\shortcite{Whyte2015}. 
% Bhandari2016: http://ieeexplore.ieee.org/stamp/stamp.jsp?arnumber=7559994
%

\subsubsection{Reconstruction of non-line-of-sight geometry}
\label{sec:nlos}
A recent analysis of the problem of non-line-of-sight (NLOS) geometry recovery and its feasibility with ToF-based devices can be found in the work of Kadambi et al.~\shortcite{Kadambi2016TOG}. Although this problem has been tackled in other imaging modalities, such as radar, a seminal paper in our area was the work of Velten et al.~\shortcite{Velten2012nc}, who provided a method to reconstruct NLOS in a controlled laboratory setup (similar to femto-photography) but without the need for scene priors, by using a backpropagation technique, and leveraging the extremely high temporal resolution of the time profiles recovered. This work was later extended by Gupta et al.~\shortcite{Gupta2012}, with the aim of improving the analysis and robustness of the reconstruction method. 

The idea that NLOS reconstruction was possible with a somehow similar setup was raised by Kirmani et al.~\shortcite{Kirmani2009ICCV,Kirmani2011}. Later, Laurenzis and Velten~\shortcite{Laurenzis2014area}, and Buttafava et al.~\shortcite{Buttafava2015} generalized Velten's backpropagation NLOS reconstruction using data captured with gated systems and SPAD respectively. 
Heide and colleagues~\shortcite{Heide2014diffuse} presented a technique with less expensive hardware based on AMCW systems (see \Sec{phase}), although requiring significant modifications to off-the-shelf hardware. Their technique works in the presence of ambient illumination, and relies on scene priors such as sparsity of the geometry to regularize the non-linear optimization problem that arises when the linear image formation model (after~\cite{Heide2013}) is inverted. Hullin~\shortcite{Hullin2014} described an analysis-by-synthesis approach: forward light transport is implemented using radiosity, and posed the problem as an optimization that deforms the geometry (starting from a tessellated spherical blob), until the error between the captured radiance and the output of the forward model is minimized. As the author notes, convexity, and thus convergence of the optimization problem to the global solution, is not guaranteed. A generative analysis-by-synthesis approach was also used by Klein et al.~\shortcite{Klein2016SR},  to detect and track NLOS objects under a number of simplifying assumptions, employing only a laser pointer and a conventional 2D camera.
%
% Kadambi2016TOG: http://web.media.mit.edu/~hangzhao/papers/occluded.pdf
% gupta2012: https://www.osapublishing.org/DirectPDFAccess/012B9823-D586-278A-987509C4ABD26855_240434/oe-20-17-19096.pdf?da=1&id=240434&seq=0&mobile=no
% Klein2016SR: http://www.nature.com/articles/srep32491

On a theoretical level, recently, Tsai et al.~\shortcite{Tsai2016} developed a framework for shape recovery for the particular case of two-bounce light paths. The problem of separating these paths specifically remains unanswered, but the work provides the foundation for future attempts. 

%Heide2014; resolution: it achieves a resolution that ranges from a few centimeters to tenths of them, depending on scene content

\subsubsection{Imaging through scattering media} 
\label{sec:scattering}
%\diegoc{not everything here is geometry. This could maybe be its own subsection?}
There are many works regarding imaging through scattering media, traditionally focusing on medical or deep-tissue imaging~\cite{Han2000}, or inspection of art~\cite{Abraham2010}, to name a few examples.
Transient data has also been recently used for this goal. Heide et al.~\shortcite{Heide2014} introduced a convolutional sparse coding approach using correlation image sensors. In particular, the authors used a modified \ToF camera, combined with a physically-based transient image formation model (see \Sec{forward_models}) that improves sparsity in scattering media. Using instead a femto-photography setup~\cite{Velten2013}, Raviv and colleagues~\shortcite{Raviv2014} obtained the six degrees of freedom (3D position and orientation) of rigid known geometric shapes, by leveraging scattering information. The authors made the observation that obtaining the full image is not always convenient or necessary; with less information to recover, their single-shot system can be used in dynamic scenes, allowing to track the object being imaged. Femto-photography was also employed by Naik et al.~\shortcite{Naik2014} to recover the spatially-varying reflectance of a scene seen through a scattering medium by means of a numerical inversion algorithm. The method requires the time spent in the scattering medium to be lower than the temporal resolution of the camera. Last, Satat and co-workers~\shortcite{Satat2015b} combined time-resolved information with an optimization framework to image a scene behind a relatively thick scattering medium of 1.5cm. 

\begin{figure}[t]
	\centering
	\includegraphics[width=\columnwidth]{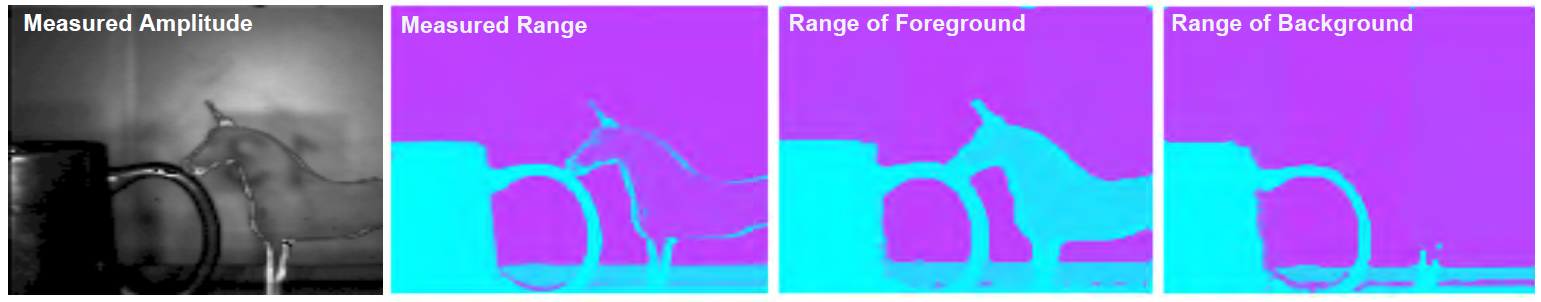} 
	\caption{\label{fig:unicorn} Given a scene with transparent objects (a), regular time of flight cameras fail at reconstructing their depth (b). Kadambi and colleagues' method, assuming that light transport is modeled as a sparse set of Dirac deltas, can correctly recover the depth of the unicorn (c and d). Figure from \protect\cite{Kadambi2013}.}
\end{figure}

Highly related is the problem of recovering the shape (depth) of transparent objects and their backgrounds. An example was shown by Kadambi et al~\shortcite{Kadambi2013} (see \Fig{unicorn}), in the context of their sparse deconvolution approach to the multipath interference problem.  Lee and Shim presented a two-step approach, using a skewed ToF stereo camera~\cite{Lee2015}. First the transparent object is detected by analyzing inconsistencies between the views from the two cameras; depth is then recovered by means of optimization, minimizing the distance between inconsistent points along their respective light rays. The same authors later proposed a second technique, this time using a single ToF camera~\cite{Shim2016}. Although lighter on the hardware side, the method requires two different measurements, with and without the transparent object. By directly analyzing the distortions created by a transparent object in ToF profiles, Tanaka and colleagues~\shortcite{Tanaka2016} showed that the refractive light path (from which depth can be inferred) can be uniquely determined with a single parameter. This is estimated with the help of a known reference board, moved to two different locations behind the transparent object.
%Naik2014: http://web.mit.edu/naik/www/assets/pdf/naik_josaa_14.pdf
%
%

%% file: src41t_table.tex
%!TEX root = src_main.tex
\begin{table*}[]
\centering
\footnotesize
\begin{tabular}{llll}
\textbf{Work}           & \textbf{Multipath Type} & \textbf{Solution Type}        & \textbf{Hardware Modifications} \\ \hline
\cite{Fuchs2010}      & Continuous              & Iterative                     & None                            \\
\cite{Dorrington2011} & 2-sparse                & Iterative                     & Frequency Sweep                 \\
\cite{Godbaz2012}     & 2-sparse                & Closed-form                   & Frequency Sweep                 \\
\cite{Kadambi2013}    & K-sparse                & Iterative                     & Custom code                     \\
\cite{Kirmani2013}    & K-sparse                & Iterative                     & Frequency Sweep                 \\
\cite{Heide2013}      & K-sparse                & Sparse Regularization                  & Frequency Sweep                 \\
\cite{Freedman2014}   & K-sparse                & Iterative                     & None                            \\
\cite{Jimenez2014}    & K-sparse                & Iterative                     & None                            \\
\cite{OToole2014}     & Continuous              & None                          & Extensive                       \\
\cite{Lin2014}        & Continuous              & Closed-form                   & Frequency Sweep                 \\
\cite{Gupta2014}      & Continuous              & Closed-form                   & Extensive                       \\
\cite{Naik2015}       & Continuous              & Closed-form                   & External Projector              \\
\cite{Peters2015}     & K-sparse                & Closed-form                   & Frequency Sweep                 \\
\cite{Qiao2015}      & K-sparse                & Sparse Regularization & Frequency Sweep                 \\
\cite{Kadambi2015}    & Continuous              & Closed-form                   & Frequency Sweep                 \\
\cite{Whyte2015}     & Continuous              & Closed-form                   & Custom code                    
\end{tabular}
\caption{Comparison between the different existing techniques addressing the problem of MPI. Adapted and updated from \protect\cite{Naik2015}.}
\label{tab:mpi}
\end{table*}

%% file: src42_motion_estimation.tex
%!TEX root = src_main.tex
\subsection{Motion estimation}
\label{sec:motion}

%A number of techniques are known that perform range imaging, including---but not limited to---: lidar, radar, stereo, coded apertures, structured light, or tomography. Among them, some, such as radar or sonar, provide the ability to perform range imaging of objects which are not in the line of sight of the camera. Transient data can help solve long-standing problems in direct line-of-sight range imaging, 
%%e.g., aiding to solve the MPI problem \belenc{can this be considered direct line-of-sight? maybe I should remove it}, 
%but the greater gains of this type of data come in the ability to observe objects that are not in the line of sight, or, as usually termed, that are ``around the corner".

Transient data also allows the detection of non-line-of-sight \textit{moving} objects. First tackled by Pandharkar et al.~\shortcite{Pandharkar2011}, %\cite{PandharkarThesis2011,Pandharkar2011,PandharkarPatent2012},
the proposed approach utilized a femtosecond laser as illumination source, and data is obtained from a streak camera at picosecond resolution, with equipment similar to the one shown in Figure~\ref{fig:femto}, but a different setup. The tracked object is located at each frame by backprojecting the recorded signals, and solving a constrained least squares problem. From these locations, the motion vector is obtained. 
%\diegoc{I commented out Klein 2016} 
%More recently, Klein et al.~\shortcite{Klein2016} proposed an analysis-by-synthesis approaching,  replacing the backprojection with an optimization step that iteratively compares the recorded wavefront with a simulated one corresponding to the estimated solution. They use cheaper hardware and indirect diffuse light reflections, and show results on two-dimensional cutouts of white, diffuse objects. 

%Their main improvement is the use of lower-cost hardware, and of intensity images only; this work thus does not make use of time of flight data, nor, strictly speaking, of transient data, since it takes as input multiple frames captured by a high-end 2D camera. 

A different approach relied on the well-known Doppler effect to obtain the velocity field of a given scene, using a \ToF camera~\cite{Heide2015}. The shift in frequency of the illumination that occurs when a large radial velocity is present breaks down the conventional ToF formulation in homodyne setups. Drawing inspiration from the communications field, the authors proposed the use of orthogonal illumination and modulation frequencies, requiring two simultaneous measurements (one homodyne, one heterodyne) to recover the velocity field. An additional, phase-offsetted homodyne measurement yields a range image as well. 

%is a similar, yet simpler, problem to the 

%\belenc{Consider having this before geometry estimation, since it is a simpler (and earlier) problem? Let's see how it builds up.} \diegoc{I think we should begin with the main app, which is geometry}
%\belenc{should we say something, here or elsewhere, about radar, sonar, and these sort of techniques? I wonder...}

%Around a corner~\cite{Pandharkar2011,Klein2016}, using Doppler~\cite{Heide2015}, 

%\diegoc{missing (not really, but maybe worth knowing about): http://rfcapture.csail.mit.edu/   and   update Klein 2016: http://www.nature.com/articles/srep32491 (although it is commented out for now)}

%% file: src43_material_estimation.tex
%!TEX root = src_main.tex
\subsection{Material estimation}
\label{sec:material}
%
%Two main trends can be identified in this area: reflectance acquisition (including non-line of sight surfaces, or the presence of turbid media), and material classification.

\begin{figure}[t]
	\centering
	\includegraphics[width=.95\columnwidth]{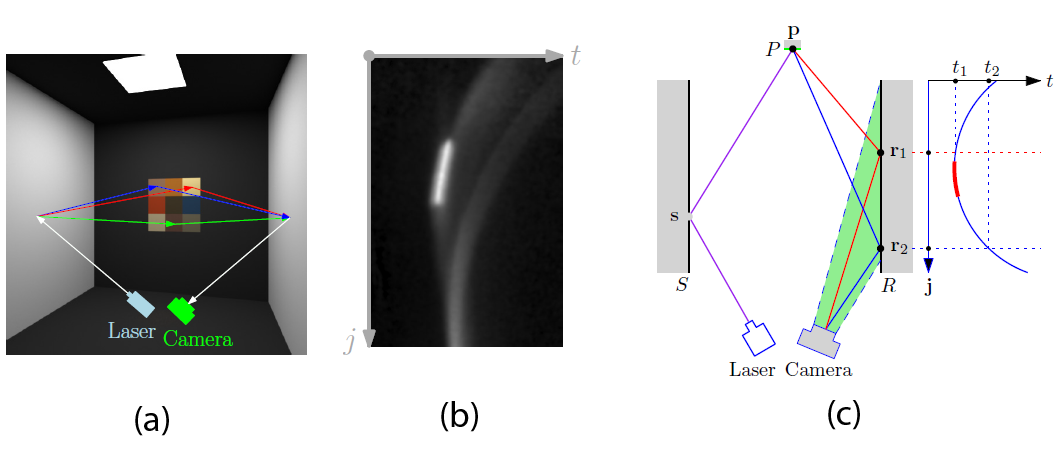}
	\caption{\label{fig:naik11} (a) Setup for reflectance acquisition using transient imaging: points in the left wall are successively illuminated, creating a range of incoming directions on the back wall; light in turn bounces off it and reflects on the right wall before reaching the camera. (b) Streak image captured with two patches on the back wall. (c) Illustration of the setup: the bright region in the streak image in (b) (red part of the rightmost graph) contains information about the shape of the specular lobe at P. Figure from \protect\cite{Naik2011}. }
\end{figure}

%\paragraph{Reflectance acquisition} 
Reflectance acquisition is traditionally a time-consuming process, due to the large number of measurements needed to fully capture a generic BRDF. Transient data, with its potential ability to disentangle light paths, has enabled alternative means of reflectance acquisition, including ``around the corner" setups. Early work in this regard was carried out by Pandharkar~\shortcite[Ch. 5]{PandharkarThesis2011}, later improved by Naik et al.~\shortcite{Naik2011}, based on a three-bounce setup with planar walls. Light from a laser bounces off a diffuse surface, then is reflected off a planar sample of the material, before reaching a third surface, also diffuse, which is observed by a streak camera. The laser needs to be swept over a number of positions on the first surface, but different outgoing directions are captured in a single shot (Figure~\ref{fig:naik11}). 
%With known planar geometry, and the ability to separate light paths, reflectance can be recovered from the recorded intensity at each point. 
Different light paths with the same path length pose a challenge; the authors formulated the reflectance recovery problem as an undetermined linear system, modeling the reflectance with a low-dimensional parametric BRDF model. Subsequent work using the same hardware was able to realize reflectance acquisition for objects behind a diffuser, which has applications in biological and medical imaging~\cite{Naik2014}. This requires estimating not only reflectance but also the scattering profile of the diffuser (assumed Gaussian). The authors formulated the forward model and then solved a convex optimization problem minimizing the difference between estimated and measured intensity.
%, which is reduced in the case of the around the corner setup.
%The multiple bounces of light enable a single-shot, direct measuerment of the BRDF of 
Tsai et al.~\shortcite{Tsai2016} described reflectance estimation by taking advantage of the coverage of different incident and outgoing ray directions given by two-bounce paths for certain scene geometries.  Different from other works, this approach did not require all surfaces to be Lambertian.  
% Material estimation~\cite{Naik2011,Naik2014,Tsai2016}.

\subsubsection{Fluorescent lifetime imaging}
The technique known as fluorescence lifetime imaging allows to obtain reflectance properties of surfaces through turbid media. It has applications in diagnosis or inspection, and is one of the most classical applications of ultrafast time-resolved imaging using impulse based illumination, with exposure times in the order of a few hundred picoseconds (see e.g. \cite{Wu1995}). Using a femtosecond laser and a streak camera, Satat et al.~\shortcite{Satat2015} utilized time-resolved images to localize and classify the lifetimes of fluorescent probes, which need not be in the line of sight of the camera. Another approach to the problem, which required simpler hardware but is only able to reconstruct directly visible fluorescent samples, is the work by Bhandari et al.~\shortcite{Bhandari2015fluor}. A ToF Kinect sensor was used to obtain lifetime information from samples, without calibration nor known illumination.  %\diegoc{the missing piece is missing: what are they sacrificing?}
%\belenc{This last work could go where we talked about fd-tof vs. td-tof. The first is related to the backproj. approaches for geometry around the corner.}

%\diegoc{I'm leaving this comment here, but wouldn't it be a shame to not cite Adrian's work with polarization? If the Chinese could put together a quick arxiv in the next couple of weeks or so... Same with Victor's work (TBD)}

\subsubsection{Material classification} 
A related but simpler problem is material classification. Wu and colleagues \shortcite{Wu2014} decomposed global light transport into its direct, subsurface scattering, and interreflection components, by analyzing the time profile at picosecond resolution. They noted that the time profile of a point lit by subsurface scattering decays exponentially, which allowed them to identify translucent materials in a scene. Further exploring a similar idea, Su et al.~\shortcite{Su2016} later analyzed four materials with different degrees of translucency, and  tested several learning methods on the captured data. The authors achieved varying success rates identifying the materials: Wax was easy to classify, due to its strong subsurface scattering, while progressively less translucent materials, such as paper, styrofoam or towel, became more complex. 
%Despite the limited range of materials that are correctly classified, it serves as a proof of concept of an alternative approach to material classification for future approaches to build upon.
%Material classification~\cite{Su2016} --- flourescent materials~\cite{Satat2015,Bhandari2015fluor}.

%% file: src50_simulation.tex
%!TEX root = src_main.tex
\section{Simulation}
\label{sec:simulation}

Light transport, described using either Maxwell's equations~\cite{Born2002}, or the more practical radiative approximation~\cite{Chandrasekhar1960}, is defined in a time-resolved manner. However, since the final goal is usually to compute light transport in steady-state, the practical assumption that the speed of light is infinite becomes a reasonable approximation from a simulation (rendering) perspective. See e.g. \cite{Gutierrez2008course,Krivanek2013} for an overview on steady-state rendering.

%In its most general form (based on Maxwell equations~\cite{Born2002}), light transport is defined in a time-resolved manner. Therefore, simulating light transport within the framework of wave-optics is done in transient state~\cite{Moravec1981,Musbach2013} \diegoc{not sure this logic makes sense; probably rewrite}. 
%
%However, this approach is not practical for applications in graphics and vision, where approaches based on radiative transfer~\cite{Chandrasekhar1960} are preferred. Thus, the vast majority of works on light transport simulation in these fields is based on geometric optics, ignoring the transient nature of light propagation (see e.g. \cite{Gutierrez2008course,Krivanek2013}). In other words, the speed of light is considered to be infinite, which is a reasonable approximation in most cases. 

With the establishment of transient imaging in graphics and vision, the simulation of time-resolved light transport is becoming an increasingly important tool. Smith et al.~\shortcite{Smith2008} developed the first framework in the context of the traditional rendering equation~\cite{Kajiya1986}. This was later formalized by Jarabo et al.~\shortcite{Jarabo2014}, extending the path integral~\cite{Veach1997Thesis} to include time-resolved effects such as propagation and scattering delays. 

\begin{figure*}[ht]
\centering
\includegraphics[width=.9\textwidth]{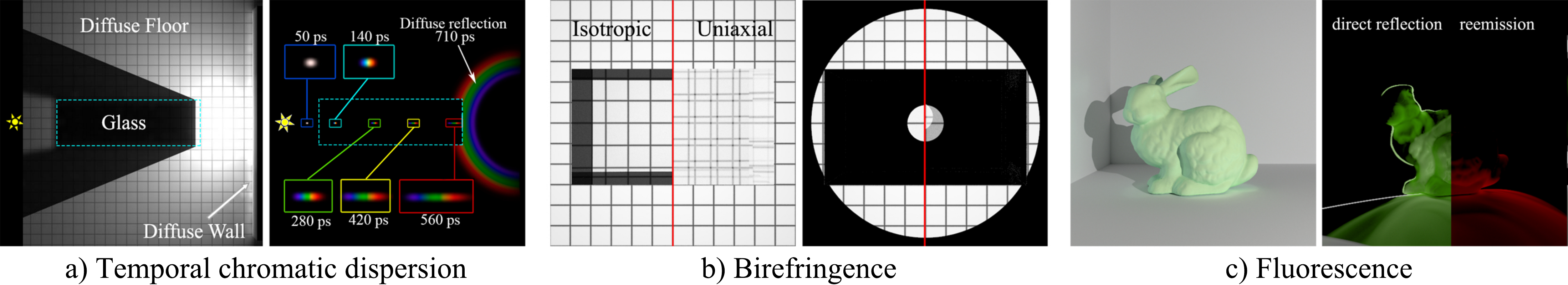}
\caption{Examples of phenomena observed in transient state: From left to right, wavelength-dependent indices of refraction produce temporal chromatic dispersion; temporal decomposition of ordinary and extraordinary transmission in a birefringent crystal; and energy re-emission after 10 nanoseconds in a fluorescent object (image from \protect \cite{Jarabo2014}).}
\label{fig:transient_effects} 
\end{figure*}

Transient rendering has been used to synthesize videos of light in motion~\cite{Jarabo2014}, but is also a key tool to provide ground truth information to develop novel light transport models~\cite{OToole2014,Adam2015}, or benchmarking~\cite{Nair2013,Pitts2014}. It can also be used as a forward model for solving inverse problems~\cite{Keller2007,Keller2009,Fuchs2008,Fuchs2010,Jimenez2012,Jimenez2014,Hullin2014,Klein2016SR}. 

The key \textbf{differences} with respect to steady-state simulation are: 
\begin{itemize}
\item The speed of light can no longer be assumed to be infinite, so propagation delays need to be taken into account. Note that some works in steady-state rendering also need to account for propagation delays (e.g. rendering based on wave-optics~\cite{Moravec1981,Musbach2013}, or solving the Eikonal equation for non-linear media~\cite{Gutierrez2005,Ihrke2007}), although their final goal is to obtain a steady-state image integrated in time.
\item Scattering causes an additional delay, due to the electromagnetic and quantum mechanisms involved in the light-matter interaction. These give rise to effects such as fluorescence, or Fresnel phase delays (see \Fig{transient_effects}).
\item The temporal domain must be reconstructed; however, naive reconstruction strategies (i.e. frame-by-frame) are extremely inefficient.
\item  Motion in the scene (e.g. camera movements) brings about the need to include relativistic effects.
\end{itemize}

%The first obvious difference between steady-state and time-resolved simulation is . Another more subtle but important difference is how the temporal domain is reconstructed. As discussed by Jarabo et al.~\shortcite{Jarabo2014}, rendering each transient frame independently is highly impractical, given the extremely short exposure times: sampling paths with a given temporal delay is almost \diegoc{almost?} impossible, while randomly sampling paths would be extremely inefficient. 

In the following, we discuss the different approaches for effectively reconstructing the temporal radiance profile in simulation; then, orthogonally to reconstruction, we focus on the different algorithms for simulating transient light transport, and on their main target application (see \Tab{render} for an overview).

\input{src50t_table}

\subsection{Reconstruction of the temporal profile}

%%RECONSTRUCCION DEL PERFIL TEMPORAL*******
 As discussed by Jarabo et al.~\shortcite{Jarabo2014}, rendering each transient frame independently is highly impractical, given the extremely short exposure times: sampling paths with a given temporal delay is almost impossible, while randomly sampling paths would be extremely inefficient.
The most straightforward way to solve this issue and render effectively transient light transport is to reuse the samples for all frames, binning them in the temporal domain~\cite{Jarabo2012,Marco2013,OToole2014,Ament2014,Adam2015,Pitts2014}. This is equivalent to a histogram density estimation; although easy to implement, it has a slow convergence of $\Order{N^{-\frac{1}{3}}}$, with $N$ being the number of samples. Jarabo et al.~\shortcite{Jarabo2014} presented a better alternative, proposing a reconstruction method based on kernel density estimation~\cite{Silverman1986}, which leads to faster convergence ($\Order{N^{-\frac{4}{5}}}$). Interestingly, rendering each frame independently, and using the histogram in the temporal domain, are equivalent to the gate imaging and streak imaging techniques discussed in \Sec{impulse}, respectively. 

If the goal is not to generate the full transient profile, but just the modulated response at the sensor as if it were captured by a correlation-based sensor (see \Sec{phase}), the problem is reduced to generating a single image modulating each sample according to its delay and the sensor response.
Thus, while we still need to keep track of the path propagation delays, it can be done within the framework of the traditional path integral, where the sensor response is a function of time. 
For depth recovery, Keller and colleagues~\shortcite{Keller2007,Keller2009} proposed a GPU-accelerated rendering system modeling such response. The system is limited to single-bounce scattering, so it assumes no MPI. 
The sensor response needs accurate sensor modulation models, including temporal behavior and noise. Gupta et al.~\shortcite{Gupta2014} introduced a noise model for AMCW imaging devices, while Lambers et al.~\shortcite{Lambers2015} presented other physically-based models of the sensor and the illumination, including high-quality noise and energy performance.

\subsection{Light transport simulation algorithms}
Depending on the application domain, existing algorithms to simulate transient light transport trade off accuracy for speed. As a forward model for efficient reconstruction of the geometry of occluded objects, Hullin~\shortcite{Hullin2014} and Klein et al.~\shortcite{Klein2016SR} extended Smith et al.'s~\shortcite{Smith2008} transient version of the radiosity method~\cite{Goral1984} on the GPU. This method is limited to Lambertian surface reflections, and second-bounce interactions. 

%GROUND TRUTH
On the other hand, most works aiming at generating \textit{ground truth} data have used transient versions of Monte Carlo (bidirectional) path tracing (BDPT)~\cite{Jarabo2012,Jarabo2014,Pitts2014,Adam2015}. These are unbiased methods, and support arbitrary scattering functions, including participating media. However, they are in general slow, requiring thousands of samples to converge. To accelerate convergence, Jarabo et al.~\shortcite{Jarabo2014} introduced three techniques for uniform sampling in the temporal domain targeted to bidirectional methods, while Lima et al.~\shortcite{Lima2011} and Periyasamy and Pramanik~\shortcite{Periyasamy2016} proposed importance sampling strategies in the context of Optical Coherence Tomography. These techniques are designed to work in the presence of participating media; this is a particularly interesting case for transient imaging, since one of its key applications is seeing through such media (fog, murky water, etc).

%METER BIAS PARA SER MAS ROBUSTO
Other algorithms aiming to produce ground truth data robustly rely on a photon tracing and gathering approach~\cite{Jensen2001,Hachisuka2013course}. Meister and colleagues~\shortcite{Meister2013,Meister2013vmv} used a transient version of photon mapping, resulting into a robust estimation of light transport, and allowing to render caustics in the transient domain. Ament et al.~\shortcite{Ament2014} also used transient photon mapping to solve the refractive RTE. However, these techniques are intrinsically biased, due to the density estimation step at the core of the photon mapping algorithm. This bias was reduced by Jarabo and colleagues~\shortcite{Jarabo2014}, who introduced progressive density estimation along the temporal domain. Targeted to transport in media, Marco~\shortcite{Marco2013} proposed a transient version of the photon beams algorithm, which was later implemented in 2D by Bitterli~\shortcite{Bitterli2016}. 

Last, as mentioned above, camera movements at this temporal resolution bring about the need to simulate relativistic effects in transient light transport. These were simulated by Jarabo and colleagues~\cite{Jarabo2013,Jarabo2015}, including time dilation, light aberration, frequency shift, radiance accumulation and distortions on the camera's field of view. The system considered linear motion, as well as acceleration and rotation of the camera.

%% file: src50t_table.tex
%!TEX root = src_main.tex
\begin{table*}[]
\centering
\footnotesize
\begin{tabular}{lllll}
\textbf{Work}                  & \textbf{Output} & \textbf{Algorithm} & \textbf{Global Illumination} & \textbf{Convergence} \\ \hline
\cite{Keller2007,Keller2009} & 4-Bucket Image  & Rasterization      & No                           & $O(1)$               \\
\cite{Jarabo2012}            & Transient Image & BDPT               & Yes                          & $O(N^{1/3})$         \\
\cite{Meister2013}           & 4-Bucket Image  & Photon Mapping     & Yes                          & $O(N^{-2/3})^*$       \\
\cite{Marco2013}             & Transient Image & Photon Beams       & Media only                   & $-$                  \\
\cite{Jarabo2014}            & Transient Image & BDPT               & Yes                          & $O(N^{-4/5})$         \\
\cite{Pitts2014}             & Transient Image & BDPT               & Yes                          & $O(N^{-1/3})$         \\
\cite{Hullin2014,Klein2016SR}  & Transient Image & Radiosity          & 2nd bounce                   & $O(1)^*$              \\
\cite{Ament2014}             & Transient Image & Photon Mapping     & Yes                          & $O(N^{-1/3})$      \\
\cite{Adam2015}              & Transient Image & BDPT / MTL         & Yes                          & $O(N^{-1/3})$         
\end{tabular}
\caption{Comparison of selected works on time-resolved rendering, including their output, the type of transport being simulated, and their convergence rate with respect to the number of samples $N$. Note that the works based on radiosity~\protect\cite{Hullin2014,Klein2016SR} (marked with an asterisk) have constant convergence, but their cost and accuracy depend on the number of geometric subdivisions in the scene. The work of Meister et al.~\protect\shortcite{Meister2013} converges with the same rate as traditional steady-state progressive photon mapping.   }
\label{tab:render}
\end{table*}

%% file: src60_discussion.tex
%!TEX root = src_main.tex
\section{Conclusions}
\label{sec:discussion}
Systems capable to obtain scene information from the temporal  response of light have existed for a while, mostly focusing on range imaging (e.g. LIDAR systems based on laser impulse or on \ToF sensors. However, the numerous problems and limitations of these sensors  (low resolution, MPI, etc.) have made them practical only in a limited set of scenarios. 
Recent advances in transient imaging and \ToF technology have, on the other hand, triggered significant improvements on the main, primal application of \ToF and LIDAR systems (i.e. range imaging), by providing more robust solutions to the MPI problem. As a result, this has expanded the range of applications and accuracy of this technology, becoming a gold standard for single image depth recovery. 
Beyond that, transient imaging has opened a vast new field of applications, making traditional and new ill-posed problems in computer vision tractable: Non-line-of-sight vision, or vision through turbid media, are some of the most exciting examples of these new applications. Other problems such as material capture and recognition, motion estimation, dehazing, or bare-sensor imaging have also been tackled successfully using transient imaging. In many cases, the combination of computer graphics, vision, and imaging techniques has played an important role, making transient imaging a truly multidisciplinary field. 

Of course, there is still much work to do: While improved hardware has now reached the market (e.g. the Kinect 2 sensor), most other applications are still restricted to controlled laboratory conditions. Limitations in hardware (poor SNR, long capture times), and software (expensive post-processing, lack of general enough priors) are still too restrictive to move these other applications (e.g. NLOS) into the wild. 
We hope that by categorizing the current state of the art of the field in a holistic way, including all the pieces we believe crucial for transient imaging (capture, simulation, analysis and applications), will help provide both a clear picture of the main  limitations of the existing pipelines, as well as the  current development of applications enabled by transient imaging.
%
%Finally, we would like to emphasize the relation of light-based transient imaging with other fields: therahertz, microwaves, radio or even sound capture and processing are also mature fields, which have tried to tackle some of the problems the vision and graphics communities aim to solve using transient imaging; we refer to additional sources (e.g. \cite{Satat2016}) to complement our work from other fields' perspective.

%% file: srcZ_acks.tex
%!TEX root = src_main.tex
\section*{Acknowledgments}
We would like to thank Matthias Hullin for discussions and comments on early versions of the work. 
This research has been partially funded by DARPA (project REVEAL), the European Research Council (Consolidator Grant, project CHAMELEON), and the Spanish Ministerio de Econom\'{i}a y Competitividad (projects TIN2016-78753-P, TIN2016-79710-P and TIN2014-61696-EXP). Julio Marco was additionally funded by a grant from the Gobierno de Arag\'{o}n.

%% file: src_main.bbl
\begin{thebibliography}{\protect\citename{K{\v{r}}iv{\'a}nek et~al\mbox{.}
  }2013}

\bibitem[\protect\citename{Abraham et~al\mbox{.} }2010]{Abraham2010}
{\sc Abraham, E., Younus, A., Delagnes, J.~C., and Mounaix, P.}
\newblock 2010.
\newblock Non-invasive investigation of art paintings by terahertz imaging.
\newblock {\em Applied Physics A 100}, 3.

\bibitem[\protect\citename{Abramson }1978]{Abramson1978}
{\sc Abramson, N.}
\newblock 1978.
\newblock Light-in-flight recording by holography.
\newblock {\em Opt. Lett. 3}, 4.

\bibitem[\protect\citename{Abramson }1983]{Abramson1983}
{\sc Abramson, N.}
\newblock 1983.
\newblock Light-in-flight recording: high-speed holographic motion pictures of
  ultrafast phenomena.
\newblock {\em Applied Optics 22}, 2.

\bibitem[\protect\citename{Adam et~al\mbox{.} }2016]{Adam2015}
{\sc Adam, A., Dann, C., Yair, O., Mazor, S., and Nowozin, S.}
\newblock 2016.
\newblock Bayesian time-of-flight for realtime shape, illumination and albedo.
\newblock {\em IEEE Trans. Pattern Analysis and Machine Intelligence\/}.

\bibitem[\protect\citename{Adams and Probert }1996]{Adams1996}
{\sc Adams, M.~D., and Probert, P.~J.}
\newblock 1996.
\newblock The interpretation of phase and intensity data from amcw light
  detection sensors for reliable ranging.
\newblock {\em The International Journal of Robotics Research 15}, 5.

\bibitem[\protect\citename{Ament et~al\mbox{.} }2014]{Ament2014}
{\sc Ament, M., Bergmann, C., and Weiskopf, D.}
\newblock 2014.
\newblock Refractive radiative transfer equation.
\newblock {\em ACM Trans. Graph. 33}, 2.

\bibitem[\protect\citename{Bamji et~al\mbox{.} }2015]{Bamji2015}
{\sc Bamji, C.~S., O'Connor, P., Elkhatib, T., Mehta, S., Thompson, B.,
  Prather, L.~A., Snow, D., Akkaya, O.~C., Daniel, A., Payne, A.~D., Perry, T.,
  Fenton, M., and Chan, V.~H.}
\newblock 2015.
\newblock A 0.13 {$\mu$}m {CMOS} {System-on-Chip} for a {512$\times$424}
  time-of-flight image sensor with multi-frequency photo-demodulation up to 130
  {MHz} and 2 {GS}/s {ADC}.
\newblock {\em IEEE Journal of Solid-State Circuits 50}, 1 (Jan).

\bibitem[\protect\citename{Bhandari and Raskar }2016]{Bhandari2016}
{\sc Bhandari, A., and Raskar, R.}
\newblock 2016.
\newblock Signal processing for time-of-flight imaging sensors.
\newblock {\em IEEE Signal Processing Magazine 33}, 5.

\bibitem[\protect\citename{Bhandari et~al\mbox{.} }2014a]{Bhandari2015}
{\sc Bhandari, A., Feigin, M., Izadi, S., Rhemann, C., Schmidt, M., and Raskar,
  R.}
\newblock 2014.
\newblock Resolving multipath interference in kinect: An inverse problem
  approach.
\newblock In {\em IEEE SENSORS}.

\bibitem[\protect\citename{Bhandari et~al\mbox{.} }2014b]{Bhandari2014}
{\sc Bhandari, A., Kadambi, A., Whyte, R., Barsi, C., Feigin, M., Dorrington,
  A., and Raskar, R.}
\newblock 2014.
\newblock Resolving multipath interference in time-of-flight imaging via
  modulation frequency diversity and sparse regularization.
\newblock {\em Opt. Lett. 39}, 6.

\bibitem[\protect\citename{Bhandari et~al\mbox{.} }2015]{Bhandari2015fluor}
{\sc Bhandari, A., Barsi, C., and Raskar, R.}
\newblock 2015.
\newblock Blind and reference-free fluorescence lifetime estimation via
  consumer time-of-flight sensors.
\newblock {\em Optica 2}, 11.

\bibitem[\protect\citename{Bitterli }2016]{Bitterli2016}
{\sc Bitterli, B.}, 2016.
\newblock Virtual femto photography.
\newblock {\scriptsize \url{https://benedikt-bitterli.me/femto.html}}.

\bibitem[\protect\citename{Born and Wolf }2002]{Born2002}
{\sc Born, M., and Wolf, E.}
\newblock 2002.
\newblock {\em Principles of Optics: Electromagnetic Theory of Propagation,
  Interference and Diffraction of Light}.
\newblock Cambridge University Press.

\bibitem[\protect\citename{Busck and Heiselberg }2004]{Busck2004}
{\sc Busck, J., and Heiselberg, H.}
\newblock 2004.
\newblock Gated viewing and high-accuracy three-dimensional laser radar.
\newblock {\em Applied Optics 43}, 24.

\bibitem[\protect\citename{Buttafava et~al\mbox{.} }2015]{Buttafava2015}
{\sc Buttafava, M., Zeman, J., Tosi, A., Eliceiri, K., and Velten, A.}
\newblock 2015.
\newblock Non-line-of-sight imaging using a time-gated single photon avalanche
  diode.
\newblock {\em Opt. Express 23}, 16.

\bibitem[\protect\citename{Chandrasekhar }1960]{Chandrasekhar1960}
{\sc Chandrasekhar, S.}
\newblock 1960.
\newblock {\em Radiative Transfer}.
\newblock Dover.

\bibitem[\protect\citename{Charbon }2007]{Charbon2007}
{\sc Charbon, E.}
\newblock 2007.
\newblock Will avalanche photodiode arrays ever reach 1 megapixel.
\newblock In {\em International Image Sensor Workshop}.

\bibitem[\protect\citename{Conroy et~al\mbox{.} }2009]{Conroy2009}
{\sc Conroy, R.~M., Dorrington, A.~A., K{\"u}nnemeyer, R., and Cree, M.~J.}
\newblock 2009.
\newblock Range imager performance comparison in homodyne and heterodyne
  operating modes.
\newblock In {\em IS\&T/SPIE Electronic Imaging}.

\bibitem[\protect\citename{Dai et~al\mbox{.} }2013]{Dai2013}
{\sc Dai, H., He, W., Miao, Z., Chen, Y., and Gu, G.}
\newblock 2013.
\newblock Three-dimensional active imaging using compressed gating.
\newblock In {\em International Symposium on Photoelectronic Detection and
  Imaging}.

\bibitem[\protect\citename{D'Eon and Irving }2011]{DEon2011}
{\sc D'Eon, E., and Irving, G.}
\newblock 2011.
\newblock A quantized-diffusion model for rendering translucent materials.
\newblock {\em ACM Trans. Graph. 30}, 4.

\bibitem[\protect\citename{Dorrington et~al\mbox{.} }2007]{Dorrington2007}
{\sc Dorrington, A.~A., Cree, M.~J., Payne, A.~D., Conroy, R.~M., and Carnegie,
  D.~A.}
\newblock 2007.
\newblock Achieving sub-millimetre precision with a solid-state full-field
  heterodyning range imaging camera.
\newblock {\em Measurement Science and Technology 18}, 9.

\bibitem[\protect\citename{Dorrington et~al\mbox{.} }2011]{Dorrington2011}
{\sc Dorrington, A.~A., Godbaz, J.~P., Cree, M.~J., Payne, A.~D., and Streeter,
  L.~V.}
\newblock 2011.
\newblock Separating true range measurements from multi-path and scattering
  interference in commercial range cameras.
\newblock In {\em IS\&T/SPIE Electronic Imaging}.

\bibitem[\protect\citename{Durand et~al\mbox{.} }2005]{Durand2005}
{\sc Durand, F., Holzschuch, N., Soler, C., Chan, E., and Sillion, F.~X.}
\newblock 2005.
\newblock A frequency analysis of light transport.
\newblock {\em ACM Trans. Graph. 24}, 3.

\bibitem[\protect\citename{Falie }2009]{Falie2009}
{\sc Falie, D.}
\newblock 2009.
\newblock Improvements of the {3D} images captured with time-of-flight cameras.
\newblock {\em arXiv preprint arXiv:0909.5656\/}.

\bibitem[\protect\citename{Freedman et~al\mbox{.} }2014]{Freedman2014}
{\sc Freedman, D., Smolin, Y., Krupka, E., Leichter, I., and Schmidt, M.}
\newblock 2014.
\newblock {SRA}: Fast removal of general multipath for {ToF} sensors.
\newblock In {\em European Conference on Computer Vision}.

\bibitem[\protect\citename{Fuchs and Hirzinger }2008]{Fuchs2008}
{\sc Fuchs, S., and Hirzinger, G.}
\newblock 2008.
\newblock Extrinsic and depth calibration of {ToF}-cameras.
\newblock In {\em IEEE Computer Vision and Pattern Recognition}.

\bibitem[\protect\citename{Fuchs }2010]{Fuchs2010}
{\sc Fuchs, S.}
\newblock 2010.
\newblock Multipath interference compensation in time-of-flight camera images.
\newblock In {\em IEEE International Conference on Pattern Recognition}.

\bibitem[\protect\citename{Gao et~al\mbox{.} }2014]{Gao2014}
{\sc Gao, L., Liang, J., Li, C., and Wang, L.~V.}
\newblock 2014.
\newblock Single-shot compressed ultrafast photography at one hundred billion
  frames per second.
\newblock {\em Nature 516}, 7529.

\bibitem[\protect\citename{Gariepy et~al\mbox{.} }2015]{Gariepy2015}
{\sc Gariepy, G., Krstaji{\'c}, N., Henderson, R., Li, C., Thomson, R.~R.,
  Buller, G.~S., Heshmat, B., Raskar, R., Leach, J., and Faccio, D.}
\newblock 2015.
\newblock Single-photon sensitive light-in-fight imaging.
\newblock {\em Nature Communications 6\/}.

\bibitem[\protect\citename{Gkioulekas et~al\mbox{.} }2015]{Gkioulekas2015}
{\sc Gkioulekas, I., Levin, A., Durand, F., and Zickler, T.}
\newblock 2015.
\newblock Micron-scale light transport decomposition using interferometry.
\newblock {\em ACM Trans. Graph. 34}, 4.

\bibitem[\protect\citename{Godbaz et~al\mbox{.} }2008]{Godbaz2008}
{\sc Godbaz, J.~P., Cree, M.~J., and Dorrington, A.~A.}
\newblock 2008.
\newblock Mixed pixel return separation for a full-field ranger.
\newblock In {\em IEEE International Conference Image and Vision Computing New
  Zealand '08}.

\bibitem[\protect\citename{Godbaz et~al\mbox{.} }2009]{Godbaz2009}
{\sc Godbaz, J.~P., Cree, M.~J., and Dorrington, A.~A.}
\newblock 2009.
\newblock Multiple return separation for a full-field ranger via continuous
  waveform modelling.
\newblock In {\em IS\&T/SPIE Electronic Imaging}.

\bibitem[\protect\citename{Godbaz et~al\mbox{.} }2010]{Godbaz2010}
{\sc Godbaz, J.~P., Cree, M.~J., and Dorrington, A.~A.}
\newblock 2010.
\newblock Extending amcw lidar depth-of-field using a coded aperture.
\newblock In {\em Asian Conference on Computer Vision 2010}.

\bibitem[\protect\citename{Godbaz et~al\mbox{.} }2012]{Godbaz2012}
{\sc Godbaz, J.~P., Cree, M.~J., and Dorrington, A.~A.}
\newblock 2012.
\newblock Closed-form inverses for the mixed pixel/multipath interference
  problem in amcw lidar.
\newblock In {\em IS\&T/SPIE Electronic Imaging}.

\bibitem[\protect\citename{Goral et~al\mbox{.} }1984]{Goral1984}
{\sc Goral, C.~M., Torrance, K.~E., Greenberg, D.~P., and Battaile, B.}
\newblock 1984.
\newblock Modeling the interaction of light between diffuse surfaces.
\newblock {\em SIGGRAPH Comput. Graph. 18}, 3.

\bibitem[\protect\citename{Gupta et~al\mbox{.} }2012]{Gupta2012}
{\sc Gupta, O., Willwacher, T., Velten, A., Veeraraghavan, A., and Raskar, R.}
\newblock 2012.
\newblock Reconstruction of hidden {3D} shapes using diffuse reflections.
\newblock {\em Opt. Express 20}, 17.

\bibitem[\protect\citename{Gupta et~al\mbox{.} }2015a]{Kadambi2015CursoICCV}
{\sc Gupta, M., Kadambi, A., Bhandari, A., and Raskar, R.}, 2015.
\newblock Computational time of flight.
\newblock {In ICCV Courses.}

\bibitem[\protect\citename{Gupta et~al\mbox{.} }2015b]{Gupta2014}
{\sc Gupta, M., Nayar, S.~K., Hullin, M.~B., and Martin, J.}
\newblock 2015.
\newblock Phasor imaging: A generalization of correlation-based time-of-flight
  imaging.
\newblock {\em ACM Trans. Graph. 34}, 5.

\bibitem[\protect\citename{Gutierrez et~al\mbox{.} }2005]{Gutierrez2005}
{\sc Gutierrez, D., Mu{\~{n}}oz, A., Anson, O., and Seron, F.}
\newblock 2005.
\newblock Non-linear volume photon mapping.
\newblock In {\em Eurographics Symposium on Rendering}.

\bibitem[\protect\citename{Gutierrez et~al\mbox{.} }2008]{Gutierrez2008course}
{\sc Gutierrez, D., Narasimhan, S.~G., Jensen, H.~W., and Jarosz, W.}
\newblock 2008.
\newblock Scattering.
\newblock In {\em ACM SIGGRAPH ASIA 2008 Courses}.

\bibitem[\protect\citename{Hachisuka et~al\mbox{.} }2013]{Hachisuka2013course}
{\sc Hachisuka, T., Jarosz, W., Georgiev, I., Kaplanyan, A., and
  Nowrouzezahrai, D.}
\newblock 2013.
\newblock State of the art in photon density estimation.
\newblock In {\em ACM SIGGRAPH ASIA 2013 Courses}.

\bibitem[\protect\citename{Hamamatsu }2012]{Hamamatsu}
{\sc Hamamatsu}, 2012.
\newblock Guide to streak cameras.
\newblock {\scriptsize
  \url{http://sales.hamamatsu.com/assets/pdf/catsandguides/e_streakh.pdf}}.

\bibitem[\protect\citename{Han et~al\mbox{.} }2000]{Han2000}
{\sc Han, P.~Y., Cho, G.~C., and Zhang, X.-C.}
\newblock 2000.
\newblock Time-domain transillumination of biological tissues with terahertz
  pulses.
\newblock {\em Opt. Lett. 25}, 4 (Feb).

\bibitem[\protect\citename{Hansard et~al\mbox{.} }2012]{Hansard2012}
{\sc Hansard, M., Lee, S., Choi, O., and Horaud, R.~P.}
\newblock 2012.
\newblock {\em Time-of-flight cameras: principles, methods and applications}.
\newblock Springer Science \& Business Media.

\bibitem[\protect\citename{Hebert and Krotkov }1992]{Hebert1992}
{\sc Hebert, M., and Krotkov, E.}
\newblock 1992.
\newblock {3D} measurements from imaging laser radars: how good are they?
\newblock {\em Image Vision Comput. 10}, 3.

\bibitem[\protect\citename{Heide et~al\mbox{.} }2013]{Heide2013}
{\sc Heide, F., Hullin, M., Gregson, J., and Heidrich, W.}
\newblock 2013.
\newblock Low-budget transient imaging using photonic mixer devices.
\newblock {\em ACM Trans. Graph. 32}, 4.

\bibitem[\protect\citename{Heide et~al\mbox{.} }2014a]{Heide2014diffuse}
{\sc Heide, F., Xiao, L., Heidrich, W., and Hullin, M.~B.}
\newblock 2014.
\newblock Diffuse mirrors: {3D} reconstruction from diffuse indirect
  illumination using inexpensive time-of-flight sensors.
\newblock In {\em IEEE Computer Vision and Pattern Recognition}.

\bibitem[\protect\citename{Heide et~al\mbox{.} }2014b]{Heide2014}
{\sc Heide, F., Xiao, L., Kolb, A., Hullin, M.~B., and Heidrich, W.}
\newblock 2014.
\newblock Imaging in scattering media using correlation image sensors and
  sparse convolutional coding.
\newblock {\em Opt. Express 22}, 21.

\bibitem[\protect\citename{Heide et~al\mbox{.} }2015]{Heide2015}
{\sc Heide, F., Heidrich, W., Hullin, M., and Wetzstein, G.}
\newblock 2015.
\newblock Doppler time-of-flight imaging.
\newblock {\em ACM Trans. Graph. 34}, 4.

\bibitem[\protect\citename{Heshmat et~al\mbox{.} }2014]{Heshmat2014}
{\sc Heshmat, B., Satat, G., Barsi, C., and Raskar, R.}
\newblock 2014.
\newblock Single-shot ultrafast imaging using parallax-free alignment with a
  tilted lenslet array.
\newblock In {\em CLEO: Science and Innovations}.

\bibitem[\protect\citename{Hu et~al\mbox{.} }2014]{Hu2014}
{\sc Hu, X., Deng, Y., Lin, X., Suo, J., Dai, Q., Barsi, C., and Raskar, R.}
\newblock 2014.
\newblock Robust and accurate transient light transport decomposition via
  convolutional sparse coding.
\newblock {\em Opt. Lett. 39}, 11.

\bibitem[\protect\citename{Huang et~al\mbox{.} }1991]{Huang1991}
{\sc Huang, D., Swanson, E.~A., Lin, C.~P., Schuman, J.~S., Stinson, W.~G.,
  Chang, W., Hee, M.~R., Flotte, T., Gregory, K., Puliafito, C.~A., and
  Fujimoto, J.~G.}
\newblock 1991.
\newblock Optical coherence tomography.
\newblock {\em Science 254}, 5035.

\bibitem[\protect\citename{Hullin }2014]{Hullin2014}
{\sc Hullin, M.~B.}
\newblock 2014.
\newblock Computational imaging of light in flight.
\newblock In {\em SPIE/COS Photonics Asia}.

\bibitem[\protect\citename{Ihrke et~al\mbox{.} }2007]{Ihrke2007}
{\sc Ihrke, I., Ziegler, G., Tevs, A., Theobalt, C., Magnor, M., and Seidel,
  H.-P.}
\newblock 2007.
\newblock Eikonal rendering: Efficient light transport in refractive objects.
\newblock {\em ACM Trans. Graph. 26}, 3.

\bibitem[\protect\citename{Jamtsho and Lichti }2010]{Jamtsho2010}
{\sc Jamtsho, S., and Lichti, D.~D.}
\newblock 2010.
\newblock Modelling scattering distortion in {3D} range camera.
\newblock {\em International Archives of Photogrammetry, Remote Sensing and
  Spatial Information Sciences 38}, 5.

\bibitem[\protect\citename{Jarabo et~al\mbox{.} }2013]{Jarabo2013}
{\sc Jarabo, A., Masia, B., Velten, A., Barsi, C., Raskar, R., and Gutierrez,
  D.}
\newblock 2013.
\newblock Rendering relativistic effects in transient imaging.
\newblock In {\em Congreso Espa\~nol de Inform\'atica Gr\'afica}.

\bibitem[\protect\citename{Jarabo et~al\mbox{.} }2014]{Jarabo2014}
{\sc Jarabo, A., Marco, J., Mu\~{n}oz, A., Buisan, R., Jarosz, W., and
  Gutierrez, D.}
\newblock 2014.
\newblock A framework for transient rendering.
\newblock {\em ACM Trans. Graph. 33}, 6.

\bibitem[\protect\citename{Jarabo et~al\mbox{.} }2015]{Jarabo2015}
{\sc Jarabo, A., Masia, B., Velten, A., Barsi, C., Raskar, R., and Gutierrez,
  D.}
\newblock 2015.
\newblock Relativistic effects for time-resolved light transport.
\newblock {\em Computer Graphics Forum 34}, 8.

\bibitem[\protect\citename{Jarabo }2012]{Jarabo2012}
{\sc Jarabo, A.}
\newblock 2012.
\newblock {\em Femto-photography: Visualizing light in motion}.
\newblock Master's thesis, Universidad de Zaragoza.

\bibitem[\protect\citename{Jensen }2001]{Jensen2001}
{\sc Jensen, H.~W.}
\newblock 2001.
\newblock {\em Realistic Image Synthesis Using Photon Mapping}.
\newblock {AK} Peters.

\bibitem[\protect\citename{Jim{\'e}nez et~al\mbox{.} }2012]{Jimenez2012}
{\sc Jim{\'e}nez, D., Pizarro, D., Mazo, M., and Palazuelos, S.}
\newblock 2012.
\newblock Modeling and correction of multipath interference in time of flight
  cameras.
\newblock In {\em IEEE Computer Vision and Pattern Recognition}.

\bibitem[\protect\citename{Jim{\'e}nez et~al\mbox{.} }2014]{Jimenez2014}
{\sc Jim{\'e}nez, D., Pizarro, D., Mazo, M., and Palazuelos, S.}
\newblock 2014.
\newblock Modeling and correction of multipath interference in time of flight
  cameras.
\newblock {\em Image Vision Comput. 32}, 1 (Jan.).

\bibitem[\protect\citename{Jongenelen et~al\mbox{.} }2010]{Jongenelen2010}
{\sc Jongenelen, A.~P., Carnegie, D.~A., Payne, A.~D., and Dorrington, A.~A.}
\newblock 2010.
\newblock Maximizing precision over extended unambiguous range for tof range
  imaging systems.
\newblock In {\em Instrumentation and Measurement Technology Conference
  (I2MTC), 2010 IEEE}.

\bibitem[\protect\citename{Kadambi et~al\mbox{.} }2013]{Kadambi2013}
{\sc Kadambi, A., Whyte, R., Bhandari, A., Streeter, L., Barsi, C., Dorrington,
  A., and Raskar, R.}
\newblock 2013.
\newblock Coded time of flight cameras: sparse deconvolution to address
  multipath interference and recover time profiles.
\newblock {\em ACM Trans. Graph. 32}, 6.

\bibitem[\protect\citename{Kadambi et~al\mbox{.} }2015]{Kadambi2015}
{\sc Kadambi, A., Taamazyan, V., Jayasuriya, S., and Raskar, R.}
\newblock 2015.
\newblock Frequency domain {ToF}: encoding object depth in modulation
  frequency.
\newblock {\em arXiv preprint arXiv:1503.01804\/}.

\bibitem[\protect\citename{Kadambi et~al\mbox{.} }2016a]{Kadambi2016b}
{\sc Kadambi, A., Schiel, J., and Raskar, R.}
\newblock 2016.
\newblock Macroscopic interferometry: Rethinking depth estimation with
  frequency-domain time-of-flight.
\newblock In {\em IEEE Computer Vision and Pattern Recognition}.

\bibitem[\protect\citename{Kadambi et~al\mbox{.} }2016b]{Kadambi2016TOG}
{\sc Kadambi, A., Zhao, H., Shi, B., and Raskar, R.}
\newblock 2016.
\newblock {Occluded Imaging with Time-of-Flight Sensors}.
\newblock {\em ACM Trans. Graph. 35}, 2 (Mar.).

\bibitem[\protect\citename{Kajiya }1986]{Kajiya1986}
{\sc Kajiya, J.~T.}
\newblock 1986.
\newblock The rendering equation.
\newblock In {\em SIGGRAPH}.

\bibitem[\protect\citename{Kavli et~al\mbox{.} }2008]{Kavli2008}
{\sc Kavli, T., Kirkhus, T., Thielemann, J.~T., and Jagielski, B.}
\newblock 2008.
\newblock Modelling and compensating measurement errors caused by scattering in
  time-of-flight cameras.
\newblock In {\em Optical Engineering+ Applications}.

\bibitem[\protect\citename{Keller and Kolb }2009]{Keller2009}
{\sc Keller, M., and Kolb, A.}
\newblock 2009.
\newblock Real-time simulation of time-of-flight sensors.
\newblock {\em Simulation Modelling Practice and Theory 17}, 5.

\bibitem[\protect\citename{Keller et~al\mbox{.} }2007]{Keller2007}
{\sc Keller, M., Orthmann, J., Kolb, A., and Peters, V.}
\newblock 2007.
\newblock A simulation framework for time-of-flight sensors.
\newblock In {\em International Symposium on Signals, Circuits and Systems
  2007}.

\bibitem[\protect\citename{Kirmani et~al\mbox{.} }2009]{Kirmani2009ICCV}
{\sc Kirmani, A., Hutchison, T., Davis, J., and Raskar, R.}
\newblock 2009.
\newblock Looking around the corner using transient imaging.
\newblock In {\em IEEE International Conference on Computer Vision}.

\bibitem[\protect\citename{Kirmani et~al\mbox{.} }2011]{Kirmani2011}
{\sc Kirmani, A., Hutchison, T., Davis, J., and Raskar, R.}
\newblock 2011.
\newblock Looking around the corner using ultrafast transient imaging.
\newblock {\em International Journal of Computer Vision 95}, 1.

\bibitem[\protect\citename{Kirmani et~al\mbox{.} }2013]{Kirmani2013}
{\sc Kirmani, A., Benedetti, A., and Chou, P.~A.}
\newblock 2013.
\newblock Spumic: Simultaneous phase unwrapping and multipath interference
  cancellation in time-of-flight cameras using spectral methods.
\newblock In {\em IEEE International Conference on Multimedia and Expo}.

\bibitem[\protect\citename{Kirmani et~al\mbox{.} }2014]{Kirmani2014}
{\sc Kirmani, A., Venkatraman, D., Shin, D., Cola{\c{c}}o, A., Wong, F.~N.,
  Shapiro, J.~H., and Goyal, V.~K.}
\newblock 2014.
\newblock First-photon imaging.
\newblock {\em Science 343}, 6166.

\bibitem[\protect\citename{Klein et~al\mbox{.} }2016]{Klein2016SR}
{\sc Klein, J., Peters, C., Mart{\'\i}n, J., Laurenzis, M., and Hullin, M.~B.}
\newblock 2016.
\newblock Tracking objects outside the line of sight using {2D} intensity
  images.
\newblock {\em Scientific Reports 6\/}.

\bibitem[\protect\citename{Kolb et~al\mbox{.} }2010]{Kolb2010}
{\sc Kolb, A., Barth, E., Koch, R., and Larsen, R.}
\newblock 2010.
\newblock Time-of-flight sensors in computer graphics.
\newblock {\em Computer Graphics Forum 29}, 1.

\bibitem[\protect\citename{K{\v{r}}iv{\'a}nek et~al\mbox{.}
  }2013]{Krivanek2013}
{\sc K{\v{r}}iv{\'a}nek, J., Georgiev, I., Kaplanyan, A., and Canada, J.}
\newblock 2013.
\newblock Recent advances in light transport simulation: Theory and practice.
\newblock In {\em ACM SIGGRAPH 2013 Courses}.

\bibitem[\protect\citename{Lambers et~al\mbox{.} }2015]{Lambers2015}
{\sc Lambers, M., Hoberg, S., and Kolb, A.}
\newblock 2015.
\newblock Simulation of time-of-flight sensors for evaluation of chip layout
  variants.
\newblock {\em IEEE Sensors Journal 15}, 7.

\bibitem[\protect\citename{Lange and Seitz }2001]{Lange2001}
{\sc Lange, R., and Seitz, P.}
\newblock 2001.
\newblock Solid-state time-of-flight range camera.
\newblock {\em IEEE Journal of quantum electronics 37}, 3.

\bibitem[\protect\citename{Lange et~al\mbox{.} }2000]{Lange2000}
{\sc Lange, R., Seitz, P., Biber, A., and Lauxtermann, S.~C.}
\newblock 2000.
\newblock Demodulation pixels in {CCD} and {CMOS} technologies for
  time-of-flight ranging.
\newblock In {\em Electronic Imaging}.

\bibitem[\protect\citename{Lau et~al\mbox{.} }2016]{Lau2016}
{\sc Lau, A.~K., Tang, A.~H., Xu, J., Wei, X., Wong, K.~K., and Tsia, K.~K.}
\newblock 2016.
\newblock Optical time stretch for high-speed and high-throughput
  imaging—from single-cell to tissue-wide scales.
\newblock {\em IEEE Journal of Selected Topics in Quantum Electronics 22}, 4.

\bibitem[\protect\citename{Laurenzis and Bacher }2011]{Laurenzis2011}
{\sc Laurenzis, M., and Bacher, E.}
\newblock 2011.
\newblock Image coding for three-dimensional range-gated imaging.
\newblock {\em Applied Optics 50}, 21.

\bibitem[\protect\citename{Laurenzis and Velten }2014]{Laurenzis2014area}
{\sc Laurenzis, M., and Velten, A.}
\newblock 2014.
\newblock Nonline-of-sight laser gated viewing of scattered photons.
\newblock {\em Opt. Eng. 53}, 2.

\bibitem[\protect\citename{Laurenzis and Woiselle }2014]{Laurenzis2014cs}
{\sc Laurenzis, M., and Woiselle, A.}
\newblock 2014.
\newblock Laser gated-viewing advanced range imaging methods using compressed
  sensing and coding of range-gates.
\newblock {\em Opt. Eng. 53}, 5.

\bibitem[\protect\citename{Laurenzis et~al\mbox{.} }2007]{Laurenzis2007}
{\sc Laurenzis, M., Christnacher, F., and Monnin, D.}
\newblock 2007.
\newblock Long-range three-dimensional active imaging with superresolution
  depth mapping.
\newblock {\em Opt. Lett. 32}, 21.

\bibitem[\protect\citename{Lee and Shim }2015]{Lee2015}
{\sc Lee, S., and Shim, H.}
\newblock 2015.
\newblock Skewed stereo time-of-flight camera for translucent object imaging.
\newblock {\em Image Vision Comput. 43\/}.

\bibitem[\protect\citename{Li et~al\mbox{.} }2012]{Li2012}
{\sc Li, L., Wu, L., Wang, X., and Dang, E.}
\newblock 2012.
\newblock Gated viewing laser imaging with compressive sensing.
\newblock {\em Applied Optics 51}, 14.

\bibitem[\protect\citename{Lima et~al\mbox{.} }2011]{Lima2011}
{\sc Lima, I.~T., Kalra, A., and Sherif, S.~S.}
\newblock 2011.
\newblock Improved importance sampling for monte carlo simulation of
  time-domain optical coherence tomography.
\newblock {\em Biomedical optics express 2}, 5.

\bibitem[\protect\citename{Lin et~al\mbox{.} }2014]{Lin2014}
{\sc Lin, J., Liu, Y., Hullin, M.~B., and Dai, Q.}
\newblock 2014.
\newblock Fourier analysis on transient imaging with a multifrequency
  time-of-flight camera.
\newblock In {\em IEEE Computer Vision and Pattern Recognition}.

\bibitem[\protect\citename{Lin et~al\mbox{.} }2016]{Lin2016}
{\sc Lin, J., Liu, Y., Suo, J., and Dai, Q.}
\newblock 2016.
\newblock Frequency-domain transient imaging.
\newblock {\em IEEE Transactions on Pattern Analysis and Machine Intelligence
  PP}, 99.

\bibitem[\protect\citename{Lindner et~al\mbox{.} }2010]{Lindner2010}
{\sc Lindner, M., Schiller, I., Kolb, A., and Koch, R.}
\newblock 2010.
\newblock Time-of-flight sensor calibration for accurate range sensing.
\newblock {\em Comput. Vis. Image Underst. 114}, 12.

\bibitem[\protect\citename{Marco }2013]{Marco2013}
{\sc Marco, J.}
\newblock 2013.
\newblock {\em Transient Light Transport in Participating Media}.
\newblock Master's thesis, Universidad de Zaragoza.

\bibitem[\protect\citename{Meister et~al\mbox{.} }2013a]{Meister2013}
{\sc Meister, S., Nair, R., J\"{a}hne, B., and Kondermann, D.}
\newblock 2013.
\newblock Photon mapping based simulation of multi-path reflection artifacts in
  time-of-flight sensors.
\newblock Tech. rep., Heidelberg Collaboratory for Image Processing.

\bibitem[\protect\citename{Meister et~al\mbox{.} }2013b]{Meister2013vmv}
{\sc Meister, S., Nair, R., and Kondermann, D.}
\newblock 2013.
\newblock Simulation of time-of-flight sensors using global illumination.
\newblock In {\em Vision, Modeling \& Visualization}.

\bibitem[\protect\citename{Moravec }1981]{Moravec1981}
{\sc Moravec, H.~P.}
\newblock 1981.
\newblock {3D} graphics and the wave theory.
\newblock {\em SIGGRAPH Comput. Graph. 15}, 3.

\bibitem[\protect\citename{Musbach et~al\mbox{.} }2013]{Musbach2013}
{\sc Musbach, A., Meyer, G.~W., Reitich, F., and Oh, S.~H.}
\newblock 2013.
\newblock Full wave modelling of light propagation and reflection.
\newblock {\em Computer Graphics Forum 32}, 6.

\bibitem[\protect\citename{Naik et~al\mbox{.} }2011]{Naik2011}
{\sc Naik, N., Zhao, S., Velten, A., Raskar, R., and Bala, K.}
\newblock 2011.
\newblock Single view reflectance capture using multiplexed scattering and
  time-of-flight imaging.
\newblock {\em ACM Trans. Graph. 30\/}.

\bibitem[\protect\citename{Naik et~al\mbox{.} }2014]{Naik2014}
{\sc Naik, N., Barsi, C., Velten, A., and Raskar, R.}
\newblock 2014.
\newblock Estimating wide-angle, spatially varying reflectance using
  time-resolved inversion of backscattered light.
\newblock {\em JOSA A 31}, 5.

\bibitem[\protect\citename{Naik et~al\mbox{.} }2015]{Naik2015}
{\sc Naik, N., Kadambi, A., Rhemann, C., Izadi, S., Raskar, R., and Bing~Kang,
  S.}
\newblock 2015.
\newblock A light transport model for mitigating multipath interference in
  time-of-flight sensors.
\newblock In {\em IEEE Computer Vision and Pattern Recognition}.

\bibitem[\protect\citename{Nair et~al\mbox{.} }2013]{Nair2013}
{\sc Nair, R., Meister, S., Lambers, M., Balda, M., Hofmann, H., Kolb, A.,
  Kondermann, D., and J{\"a}hne, B.}
\newblock 2013.
\newblock Ground truth for evaluating time of flight imaging.
\newblock In {\em Time-of-Flight and Depth Imaging. Sensors, Algorithms, and
  Applications}.

\bibitem[\protect\citename{Nakagawa et~al\mbox{.} }2014]{Nakagawa2014}
{\sc Nakagawa, K., Iwasaki, A., Oishi, Y., Horisaki, R., Tsukamoto, A.,
  Nakamura, A., Hirosawa, K., Liao, H., Ushida, T., Goda, K., et~al.}
\newblock 2014.
\newblock Sequentially timed all-optical mapping photography ({STAMP}).
\newblock {\em Nature Photonics 8}, 9.

\bibitem[\protect\citename{Nayar et~al\mbox{.} }2006]{Nayar2006}
{\sc Nayar, S.~K., Krishnan, G., Grossberg, M.~D., and Raskar, R.}
\newblock 2006.
\newblock Fast separation of direct and global components of a scene using high
  frequency illumination.
\newblock {\em ACM Trans. Graph. 25}, 3.

\bibitem[\protect\citename{Ng et~al\mbox{.} }2003]{Ng2003}
{\sc Ng, R., Ramamoorthi, R., and Hanrahan, P.}
\newblock 2003.
\newblock All-frequency shadows using non-linear wavelet lighting
  approximation.
\newblock {\em ACM Trans. Graph. 22}, 3.

\bibitem[\protect\citename{O'Toole et~al\mbox{.} }2012]{OToole2012}
{\sc O'Toole, M., Raskar, R., and Kutulakos, K.~N.}
\newblock 2012.
\newblock Primal-dual coding to probe light transport.
\newblock {\em ACM Trans. Graph. 31}, 4.

\bibitem[\protect\citename{O'Toole et~al\mbox{.} }2014]{OToole2014}
{\sc O'Toole, M., Heide, F., Xiao, L., Hullin, M.~B., Heidrich, W., and
  Kutulakos, K.~N.}
\newblock 2014.
\newblock Temporal frequency probing for {5D} transient analysis of global
  light transport.
\newblock {\em ACM Trans. Graph. 33}, 4.

\bibitem[\protect\citename{Pandharkar et~al\mbox{.} }2011]{Pandharkar2011}
{\sc Pandharkar, R., Velten, A., Bardagjy, A., Lawson, E., Bawendi, M., and
  Raskar, R.}
\newblock 2011.
\newblock Estimating motion and size of moving non-line-of-sight objects in
  cluttered environments.
\newblock In {\em IEEE Computer Vision and Pattern Recognition}.

\bibitem[\protect\citename{Pandharkar }2011]{PandharkarThesis2011}
{\sc Pandharkar, R.}
\newblock 2011.
\newblock {\em Hidden object doppler: estimating motion, size and material
  properties of moving non-line-of-sight objects in cluttered environments}.
\newblock PhD thesis, Massachusetts Institute of Technology.

\bibitem[\protect\citename{Periyasamy and Pramanik }2016]{Periyasamy2016}
{\sc Periyasamy, V., and Pramanik, M.}
\newblock 2016.
\newblock Importance sampling-based monte carlo simulation of time-domain
  optical coherence tomography with embedded objects.
\newblock {\em Applied Optics 55}, 11.

\bibitem[\protect\citename{Peters et~al\mbox{.} }2015]{Peters2015}
{\sc Peters, C., Klein, J., Hullin, M.~B., and Klein, R.}
\newblock 2015.
\newblock Solving trigonometric moment problems for fast transient imaging.
\newblock {\em ACM Trans. Graph. 34}, 6.

\bibitem[\protect\citename{Pitts et~al\mbox{.} }2014]{Pitts2014}
{\sc Pitts, P., Benedetti, A., Slaney, M., and Chou, P.}
\newblock 2014.
\newblock Time of flight tracer.
\newblock Tech. rep., Microsoft.

\bibitem[\protect\citename{Qiao et~al\mbox{.} }2015]{Qiao2015}
{\sc Qiao, H., Lin, J., Liu, Y., Hullin, M.~B., and Dai, Q.}
\newblock 2015.
\newblock Resolving transient time profile in {ToF} imaging via log-sum sparse
  regularization.
\newblock {\em Opt. Lett. 40}, 6.

\bibitem[\protect\citename{Raskar and Davis }2008]{Raskar2008}
{\sc Raskar, R., and Davis, J.}
\newblock 2008.
\newblock 5d time-light transport matrix: What can we reason about scene
  properties?
\newblock Tech. rep., MIT.

\bibitem[\protect\citename{Raviv et~al\mbox{.} }2014]{Raviv2014}
{\sc Raviv, D., Barsi, C., Naik, N., Feigin, M., and Raskar, R.}
\newblock 2014.
\newblock Pose estimation using time-resolved inversion of diffuse light.
\newblock {\em Opt. Express 22}, 17.

\bibitem[\protect\citename{Remondino and Stoppa }2013]{Remondino2013}
{\sc Remondino, F., and Stoppa, D.}
\newblock 2013.
\newblock {\em {TOF Range-Imaging Cameras}}.
\newblock Springer.

\bibitem[\protect\citename{Satat et~al\mbox{.} }2015a]{Satat2015}
{\sc Satat, G., Heshmat, B., Barsi, C., Raviv, D., Chen, O., Bawendi, M.~G.,
  and Raskar, R.}
\newblock 2015.
\newblock Locating and classifying fluorescent tags behind turbid layers using
  time-resolved inversion.
\newblock {\em Nature Communications 6\/}.

\bibitem[\protect\citename{Satat et~al\mbox{.} }2015b]{Satat2015b}
{\sc Satat, G., Raviv, D., Heshmat, B., and Raskar, R.}
\newblock 2015.
\newblock Imaging through thick turbid medium using time-resolved measurement.
\newblock In {\em Imaging and Applied Optics 2015}.

\bibitem[\protect\citename{Satat et~al\mbox{.} }2016]{Satat2016}
{\sc Satat, G., Heshmat, B., Naik, N., Redo-Sanchez, A., and Raskar, R.}
\newblock 2016.
\newblock Advances in ultrafast optics and imaging applications.
\newblock In {\em SPIE Defense+ Security}.

\bibitem[\protect\citename{Sch{\"a}fer et~al\mbox{.} }2014]{Schaefer2014}
{\sc Sch{\"a}fer, H., Lenzen, F., and Garbe, C.~S.}
\newblock 2014.
\newblock Model based scattering correction in time-of-flight cameras.
\newblock {\em Opt. Express 22}, 24.

\bibitem[\protect\citename{Schwarte et~al\mbox{.} }1997]{Schwarte1997}
{\sc Schwarte, R., Xu, Z., Heinol, H.-G., Olk, J., Klein, R., Buxbaum, B.,
  Fischer, H., and Schulte, J.}
\newblock 1997.
\newblock New electro-optical mixing and correlating sensor: facilities and
  applications of the photonic mixer device ({PMD}).
\newblock In {\em Lasers and Optics in Manufacturing III}.

\bibitem[\protect\citename{Shim and Lee }2016]{Shim2016}
{\sc Shim, H., and Lee, S.}
\newblock 2016.
\newblock Recovering translucent objects using a single time-of-flight depth
  camera.
\newblock {\em IEEE Transactions on Circuits and Systems for Video Technology
  26}, 5.

\bibitem[\protect\citename{Shrestha et~al\mbox{.} }2016]{Shrestha2016}
{\sc Shrestha, S., Heide, F., Heidrich, W., and Wetzstein, G.}
\newblock 2016.
\newblock Computational imaging with multi-camera time-of-flight systems.
\newblock {\em ACM Trans. Graph. 35}, 4.

\bibitem[\protect\citename{Silverman }1986]{Silverman1986}
{\sc Silverman, B.~W.}
\newblock 1986.
\newblock {\em Density Estimation for Statistics and Data Analysis}.
\newblock Taylor \& Francis.

\bibitem[\protect\citename{Smith et~al\mbox{.} }2008]{Smith2008}
{\sc Smith, A., Skorupski, J., and Davis, J.}
\newblock 2008.
\newblock Transient rendering.
\newblock Tech. Rep. UCSC-SOE-08-26, School of Engineering, University of
  California, Santa Cruz.

\bibitem[\protect\citename{Su et~al\mbox{.} }2016]{Su2016}
{\sc Su, S., Heide, F., Swanson, R., Klein, J., Callenberg, C., Hullin, M., and
  Heidrich, W.}
\newblock 2016.
\newblock Material classification using raw time-of-flight measurements.
\newblock In {\em IEEE Computer Vision and Pattern Recognition}.

\bibitem[\protect\citename{Tadano et~al\mbox{.} }2015]{Tadano2015}
{\sc Tadano, R., Pediredla, A.~K., Mitra, K., and Veeraraghavan, A.}
\newblock 2015.
\newblock Spatial phase-sweep: Increasing temporal resolution of transient
  imaging using a light source array.
\newblock {\em arXiv preprint arXiv:1512.06539\/}.

\bibitem[\protect\citename{Tanaka et~al\mbox{.} }2016]{Tanaka2016}
{\sc Tanaka, K., Mukaigawa, Y., Kubo, H., Matsushita, Y., and Yagi, Y.}
\newblock 2016.
\newblock Recovering transparent shape from time-of-flight distortion.
\newblock In {\em IEEE Computer Vision and Pattern Recognition}.

\bibitem[\protect\citename{Tsagkatakis et~al\mbox{.} }2012]{Tsagkatakis2012}
{\sc Tsagkatakis, G., Woiselle, A., Tzagkarakis, G., Bousquet, M., Starck,
  J.-L., and Tsakalides, P.}
\newblock 2012.
\newblock Active range imaging via random gating.
\newblock In {\em SPIE Security+ Defence}.

\bibitem[\protect\citename{Tsagkatakis et~al\mbox{.} }2013]{Tsagkatakis2013}
{\sc Tsagkatakis, G., Woiselle, A., Tzagkarakis, G., Bousquet, M., Starck,
  J.~L., and Tsakalides, P.}
\newblock 2013.
\newblock Compressed gated range sensing.
\newblock In {\em SPIE Optical Engineering+ Applications}.

\bibitem[\protect\citename{Tsagkatakis et~al\mbox{.} }2015]{Tsagkatakis2015}
{\sc Tsagkatakis, G., Woiselle, A., Tzagkarakis, G., Bousquet, M., Starck,
  J.-L., and Tsakalides, P.}
\newblock 2015.
\newblock Multireturn compressed gated range imaging.
\newblock {\em Opt. Eng. 54}, 3.

\bibitem[\protect\citename{Tsai et~al\mbox{.} }2016]{Tsai2016}
{\sc Tsai, C.-Y., Veeraraghavan, A., and Sankaranarayanan, A.~C.}
\newblock 2016.
\newblock Shape and reflectance from two-bounce light transients.
\newblock In {\em IEEE International Conference on Computational Photography}.

\bibitem[\protect\citename{Veach }1997]{Veach1997Thesis}
{\sc Veach, E.}
\newblock 1997.
\newblock {\em Robust {M}onte {C}arlo methods for light transport simulation}.
\newblock PhD thesis, Stanford.

\bibitem[\protect\citename{Velten et~al\mbox{.} }2012a]{Velten2012nc}
{\sc Velten, A., Willwacher, T., Gupta, O., Veeraraghavan, A., Bawendi, M.~G.,
  and Raskar, R.}
\newblock 2012.
\newblock Recovering three-dimensional shape around a corner using ultrafast
  time-of-flight imaging.
\newblock {\em Nature Communications}, 3.

\bibitem[\protect\citename{Velten et~al\mbox{.} }2012b]{Velten2012sig}
{\sc Velten, A., Wu, D., Jarabo, A., Masia, B., Barsi, C., Lawson, E., Joshi,
  C., Gutierrez, D., Bawendi, M.~G., and Raskar, R.}
\newblock 2012.
\newblock Relativistic ultrafast rendering using time-of-flight imaging.
\newblock In {\em ACM SIGGRAPH 2012 Talks}.

\bibitem[\protect\citename{Velten et~al\mbox{.} }2013]{Velten2013}
{\sc Velten, A., Wu, D., Jarabo, A., Masia, B., Barsi, C., Joshi, C., Lawson,
  E., Bawendi, M., Gutierrez, D., and Raskar, R.}
\newblock 2013.
\newblock Femto-photography: Capturing and visualizing the propagation of
  light.
\newblock {\em ACM Trans. Graph. 32}, 4.

\bibitem[\protect\citename{Velten et~al\mbox{.} }2016]{Velten2016}
{\sc Velten, A., Wu, D., Masia, B., Jarabo, A., Barsi, C., Joshi, C., Lawson,
  E., Bawendi, M., Gutierrez, D., and Raskar, R.}
\newblock 2016.
\newblock Transient imaging of macroscopic scenes at picosecond resolution.
\newblock {\em Communications of the ACM}, to appear.

\bibitem[\protect\citename{Whyte et~al\mbox{.} }2015]{Whyte2015}
{\sc Whyte, R., Streeter, L., Cree, M.~J., and Dorrington, A.~A.}
\newblock 2015.
\newblock Resolving multiple propagation paths in time of flight range cameras
  using direct and global separation methods.
\newblock {\em Opt. Eng. 54}, 11.

\bibitem[\protect\citename{Wu et~al\mbox{.} }1995]{Wu1995}
{\sc Wu, J., Wang, Y., Perelman, L., Itzkan, I., Dasari, R.~R., and Feld,
  M.~S.}
\newblock 1995.
\newblock Time-resolved multichannel imaging of fluorescent objects embedded in
  turbid media.
\newblock {\em Opt. Lett. 20}, 5.

\bibitem[\protect\citename{Wu et~al\mbox{.} }2012a]{Wu2012cvpr}
{\sc Wu, D., O'Toole, M., Velten, A., Agrawal, A., and Raskar, R.}
\newblock 2012.
\newblock Decomposing global light transport using time of flight imaging.
\newblock In {\em IEEE Computer Vision and Pattern Recognition}.

\bibitem[\protect\citename{Wu et~al\mbox{.} }2012b]{Wu2012eccv}
{\sc Wu, D., Wetzstein, G., Barsi, C., Willwacher, T., O'Toole, M., Naik, N.,
  Dai, Q., Kutulakos, K., and Raskar, R.}
\newblock 2012.
\newblock Frequency analysis of transient light transport with applications in
  bare sensor imaging.
\newblock In {\em European Conference on Computer Vision}.

\bibitem[\protect\citename{Wu et~al\mbox{.} }2014]{Wu2014}
{\sc Wu, D., Velten, A., O'Toole, M., Masia, B., Agrawal, A., Dai, Q., and
  Raskar, R.}
\newblock 2014.
\newblock Decomposing global light transport using time of flight imaging.
\newblock {\em International Journal of Computer Vision 107}, 2.

\bibitem[\protect\citename{Xiao et~al\mbox{.} }2015]{Xiao2015}
{\sc Xiao, L., Heide, F., O'Toole, M., Kolb, A., Hullin, M.~B., Kutulakos, K.,
  and Heidrich, W.}
\newblock 2015.
\newblock Defocus deblurring and superresolution for time-of-flight depth
  cameras.
\newblock In {\em IEEE Computer Vision and Pattern Recognition}.

\bibitem[\protect\citename{Zhang and Yan }2011]{Zhang2011}
{\sc Zhang, X., and Yan, H.}
\newblock 2011.
\newblock Three-dimensional active imaging with maximum depth range.
\newblock {\em Applied Optics 50}, 12.

\bibitem[\protect\citename{Zhang et~al\mbox{.} }2012]{Zhang2012}
{\sc Zhang, X., Yan, H., and Lv, J.}
\newblock 2012.
\newblock Multireturn three-dimensional active imaging based on compressive
  sensing.
\newblock {\em Opt. Lett. 37}, 23.

\end{thebibliography}
